\documentclass[sigconf,nonacm]{aamas}

\usepackage{balance} 
\usepackage{algorithm,algorithmic}

\usepackage{amsmath}

\usepackage{xcolor}

\usepackage{multirow}
\usepackage{multicol}

\usepackage{array}  
\usepackage{caption} 
\usepackage{graphicx}

\title[PRISMATIC]{\underline{\texttt{PRISMATIC}}\includegraphics[width=1.2em,height=1.2em]{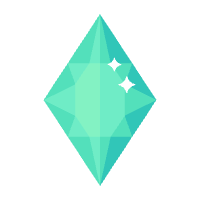}: \underline{P}rescription \underline{R}isk \underline{I}nspection \underline{S}ystem for \underline{M}ulti-\underline{A}gent \underline{T}reatment \underline{I}nteraction in \underline{C}linical Decision-Making}
\author{Ziyang Ling}
\email{zaneling@mail.ustc.edu.cn}
\affiliation{\institution{University of Science and Technology of China}\country{Hefei, China}}
\author{Shuyu Lu}
\email{Shuyu.Lu@uth.tmc.edu}
\affiliation{\institution{The University of Texas Health Science Center at Houston}\country{Houston, United States}}
\author{Ziyi Chen}
\email{chenziyi@mail.ustc.edu.cn}
\affiliation{\institution{University of Science and Technology of China}\country{Hefei, China}}

\begin{abstract}
Medical prescribing errors for multimorbid patients, who require complex drug regimens, are a major patient safety concern. These errors often stem from fragmented decision-making across medical specialties—a critical coordination failure that leads to adverse drug events and rehospitalizations.

To address this coordination challenge, we introduce \textbf{\texttt{PRISMATIC}}, a multi-agent architecture where specialized agents, each representing a workflow role, collaboratively generate and refine treatment plans. Operating on a blackboard architecture, the system first structures patient information from clinical notes, then iteratively develops safe, rational, and interpretable drug regimens. The agents' collaboration is governed by two novel adaptive coordination mechanisms: (1) Dynamic Self-Updating Weight Adjustment (DSWA), a long-term mechanism that adjusts the influence weights of agents based on historical case outcomes, enabling the system to adapt and trust more reliable specialists over time. (2) Discrepancy Feedback Precise Calibration (DFPC), a short-term mechanism where agents provide iterative feedback to resolve disagreements and eliminate prescription conflicts for the current patient.

Experiments on patient records from the MIMIC-IV database show that \textbf{\texttt{PRISMATIC}} significantly reduces drug conflicts, improves prescription coverage, and enhances treatment consistency compared to strong LLM baselines. Our work demonstrates that structuring clinical co-treatment as a multi-agent coordination problem offers a promising path toward safer and more coherent multi-disease management.
\end{abstract}


\newcommand{\BibTeX}{\rm B\kern-.05em{\sc i\kern-.025em b}\kern-.08em\TeX}
\begin{document}

\pagestyle{fancy}
\fancyhead{}

\maketitle 


\section{Introduction}
\label{sec:intro}
\begin{figure}[h]
  \centering
  \includegraphics[width=1\linewidth]{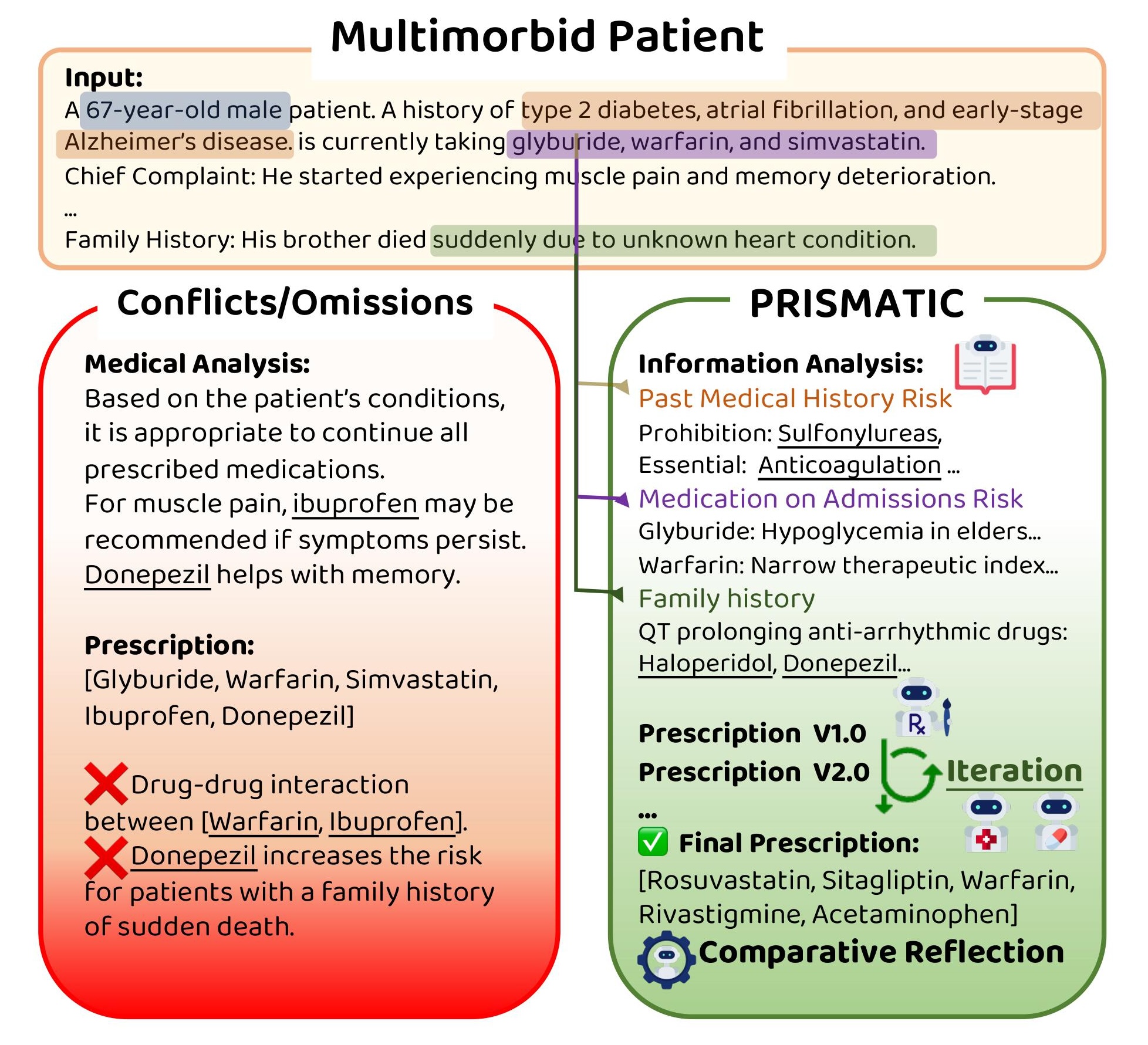}
  \caption{For patients with complex conditions and multiple comorbidities, conflicts and omissions may arise due to drug interactions and physician cognitive load. In contrast, \textbf{\texttt{PRISMATIC}} is a safe, comprehensive, and rational clinical prescription decision support system application.}
  \label{fig:logo}
  \Description{LLM vs. \textbf{\texttt{PRISMATIC}} Prescription}
\end{figure}
\begin{figure}[h]
  \centering
  \includegraphics[width=0.8\linewidth]{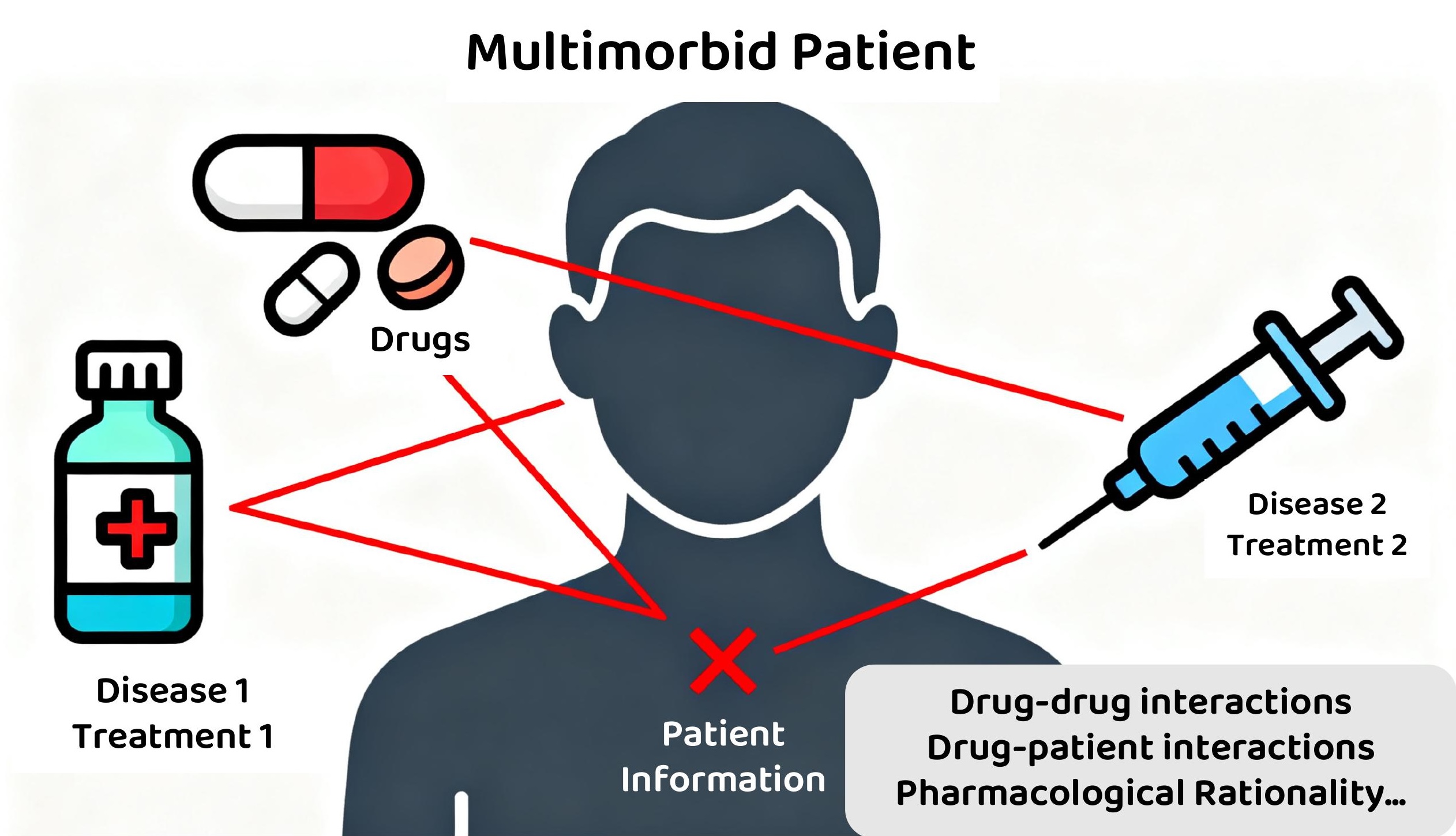}
  \caption{Multimorbid patients often experience medication conflicts and pharmacological omissions in real-world prescribing due to fragmented care and incomplete analysis of complex information.}
  \label{fig:mm}
  \Description{multimorbid patient}
\end{figure}

Prescribing for multimorbid patients: individuals with multiple chronic diseases such as hypertension, diabetes, and chronic kidney disease, is a high-stakes challenge in modern medicine \citep{ng2018patterns}. These patients require complex medication regimens often overseen by multiple specialists, yet each clinician typically focuses on a single disease-specific guideline. This siloed approach, illustrated in Figure~\ref{fig:mm}, inevitably fragments the clinical reasoning process, leading to a cascade of preventable issues: overlooked drug-drug interactions, conflicting therapeutic goals, and ultimately, adverse events. The scale of this problem is significant, with studies estimating thousands of such errors per 100,000 admissions \citep{Camacho2024}. At its core, this is not merely a problem of individual mistakes, but a systemic failure of coordination among well-intentioned but isolated decision-makers. \\

\textbf{Motivations: NLP/LLM still fails.} Early prescribing support tools relied on static rules (e.g., drug-drug interactions, contraindications) or manual chart reviews \citep{SeguraBedmar2010DDI, SeguraBedmar2011RuleBased}. Recent NLP and deep learning models automate tasks like ADE detection and medication extraction \citep{Siegersma2022ADR, Mashima2022}. Besides, while the advent of Large Language Models (LLMs) represents a significant leap in reasoning capabilities for healthcare \citep{Thirunavukarasu2023}, current applications to prescribing reveal a fundamental architectural mismatch. They typically frame the automated prescription task as a monolithic, end-to-end generation problem, ingesting all patient data to produce an output in an opaque, "black box" fashion \citep{kim2024mdagentsadaptivecollaborationllms, Chen2025, Pan2025}. This monolithic approach is fundamentally misaligned with the nature of collaborative clinical reasoning. It lacks the modularity to represent specialized knowledge domains (e.g., a clinician vs. a pharmacologist), the transparency to trace how conflicting risks were weighed, and the explicit coordination mechanisms to resolve disagreements between therapeutic goals. Ultimately, the challenge is not simply one of text generation, but of collaborative synthesis, where a task the monolithic models are ill-suited. \\

\textbf{Contributions.}
To address the gap left by monolithic models, we argue for a paradigm shift from end-to-end generation to collaborative multi-agent synthesis. In this paper, we develop this paradigm by introducing \textbf{\texttt{PRISMATIC}}, a multi-agent architecture designed to navigate the complexities of multimorbid prescribing by explicitly modeling the distributed reasoning process of the clinical scenario.\\
This paper makes the following primary contributions:
\begin{enumerate}
    \item \textbf{A Novel Multi-Agent Framework for Prescription Generation.} We design and implement \textbf{\texttt{PRISMATIC}}, a complete system that generates safer and more coherent prescriptions by proactively synthesizing fragmented clinical data and resolving conflicts.
    \item \textbf{Two Adaptive Agent Coordination Mechanisms.} We propose \textbf{DSWA} (Dynamic Self-updating Weight Adjustment) for long-term organizational learning and \textbf{DFPC} (Discrepancy Feedback Precise Calibration) for short-term, case-specific conflict resolution.
    \item \textbf{Comprehensive Empirical Validation.} We conduct extensive experiments on the combined \textbf{MIMIC-IV Note} \citep{Johnson2023MIMICIVNote} and \textbf{MIMIC-IV Hosp} \citep{johnson2024mimiciv} datasets, demonstrating that \textbf{\texttt{PRISMATIC}} significantly outperforms strong LLM and fine-tuned baselines in reducing drug conflicts, improving therapeutic coverage, and ensuring treatment consistency.
\end{enumerate}

\section{Related Works}
\textbf{Multi-Agent System in Medications.}
Multi-agent systems have long been explored in healthcare due to their decentralized, modular nature, which enables distribution of specialized tasks and supports dynamic decision-making in complex clinical settings. \\
As one of the early explorations nearly two decades ago, \citep{Rodr} proposed a rule-based agent framework to support doctor–patient collaboration and personalized hospital assistance. A decade later, \citep{BENHAJJI2015710} introduced a multi-agent system for managing patient flow and hospital resource allocation.\\
More recently, with the rapid advancement of AI and large language models (LLMs),multi-agent systems have evolved significantly, overcoming prior limitations in perception and interaction \citep{Li2024}. Recent systems leverage LLMs to enhance clinical decision-making \citep{chen2025mapevaluationmultiagentenhancement}, support surgical workflows with chain-of-thought reasoning \citep{low2025surgrawmultiagentworkflowchainofthought}, and enable collaborative diagnostic reasoning among doctor agents \citep{Chen2025}. Others incorporate verified knowledge tools \citep{gao2025txagent} or adaptive frameworks mimicking real-world clinical decision-making \citep{kim2024mdagentsadaptivecollaborationllms}. These developments underscore the growing sophistication and promise of LLM-powered multi-agent systems in improving healthcare delivery. \\

\begin{table*}[htbp]
    \centering
    \caption{The \texttt{PRISMATIC} Agent Society: Roles and Responsibilities in the Coordination Process}
    \begin{tabular}{@{} >{\centering\arraybackslash}m{3cm} | >{\centering\arraybackslash}m{3cm} | m{8.8cm} @{}}
        \hline\hline 
        \rule{0pt}{3ex}
        \textbf{Agent} & \textbf{Core Responsibility} & \textbf{Key Function in the Coordination Process} \\[1ex]
        \hline 
        
        \includegraphics[width=1em]{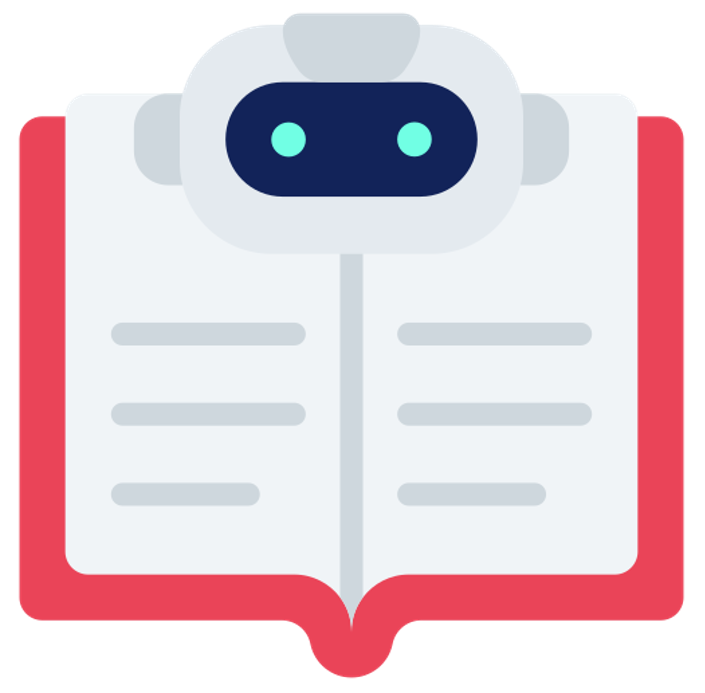} \textbf{Information Analysis (IA)} 
        & \textbf{Structure the Input} 
        & Establishes the common ground for collaboration by parsing raw clinical text (T) into a structured, shared representation (Chief Complaint (\textbf{CP}), Patient Information (\textbf{PI}) on the blackboard.
        \\[1ex] 
        \hline
        \includegraphics[width=1.6em]{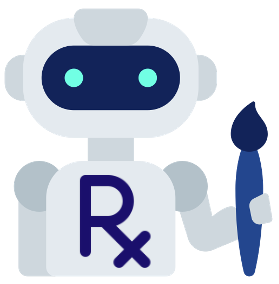} \textbf{Prescription Generation (PG)} 
        & \textbf{Propose a Solution} 
        & Acts as the primary reasoner. Synthesizes the structured information \textbf{\{CP, PI, OI\}} and contextual information from the blackboard to generate an initial candidate prescription ($\textit{P}_n$). The prescription is then \textbf{iteratively negotiated} with the \textbf{SC, PR,} and \textbf{EA} agents through the inner loop (\textbf{DFPC}), undergoing refinement based on their feedback until a consensus is reached.
        \\[1ex]
        \hline
        \includegraphics[width=1.6em]{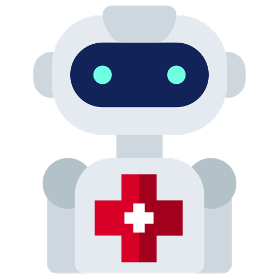} \textbf{Safety Check (SC)} 
        & \textbf{Critique on Safety} 
        & Functions as a specialized critic targeting the Safety Objective. Verifies the candidate prescription against formal safety constraints (\textbf{DDI, DPI, Dosage})\footnotemark[1] and posts specific, actionable violations for collective resolution.
        \\[1ex]
        \hline
        \includegraphics[width=1.6em]{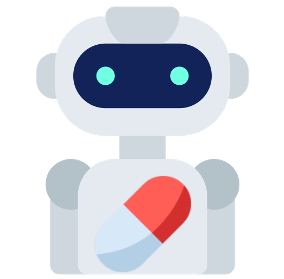} \textbf{Pharm. Rationality (PR)} 
        & \textbf{Critique on Rationality}
        & Functions as a specialized critic targeting the Rationality Objective. Evaluates the prescription's therapeutic \textbf{Coverage and Duplication}\footnotemark[2] to ensure clinical efficacy and efficiency, posting its findings as feedback.
        \\[1ex]
        \hline
        \includegraphics[width=1.6em]{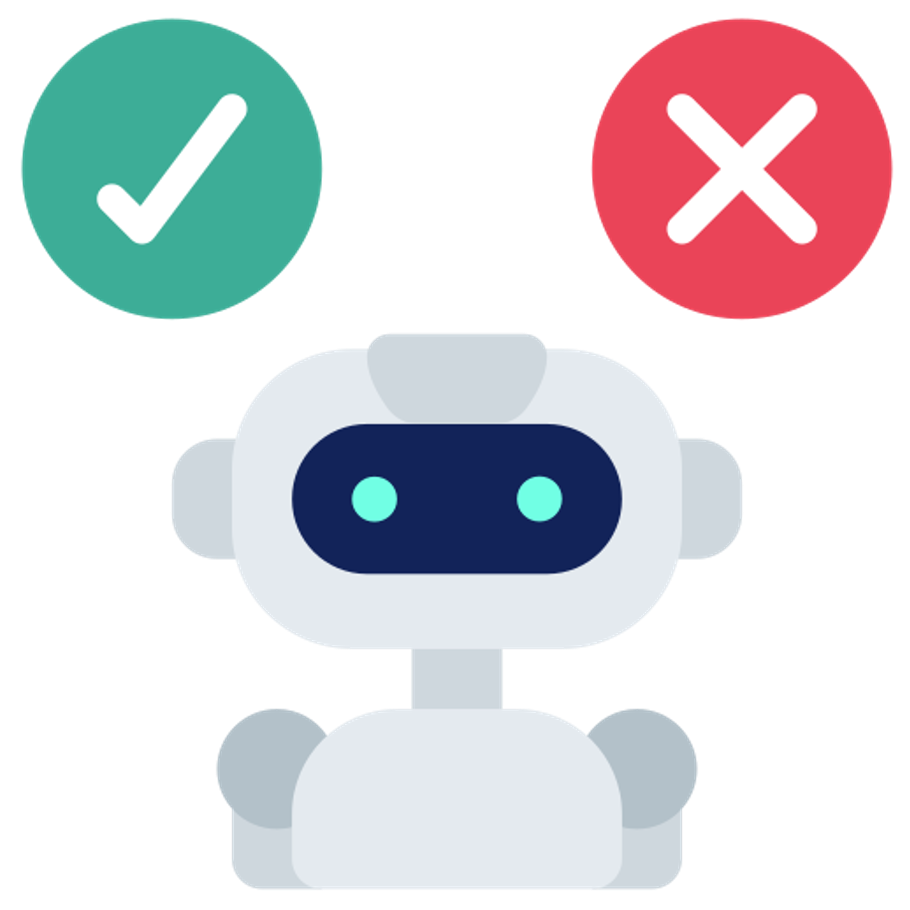} \textbf{Evaluation Agent (EA)}
        & \textbf{Adjudicate \& Decide}
        & Serves as the coordinator and adjudicator of the inner loop. It aggregates critiques from \textbf{SC} and \textbf{PR}, assesses interpretability, and computes a unified score to decide whether to accept the prescription or trigger a refinement cycle via \textbf{DFPC}. 
        \\[1ex]
        \hline
        \includegraphics[width=1.6em]{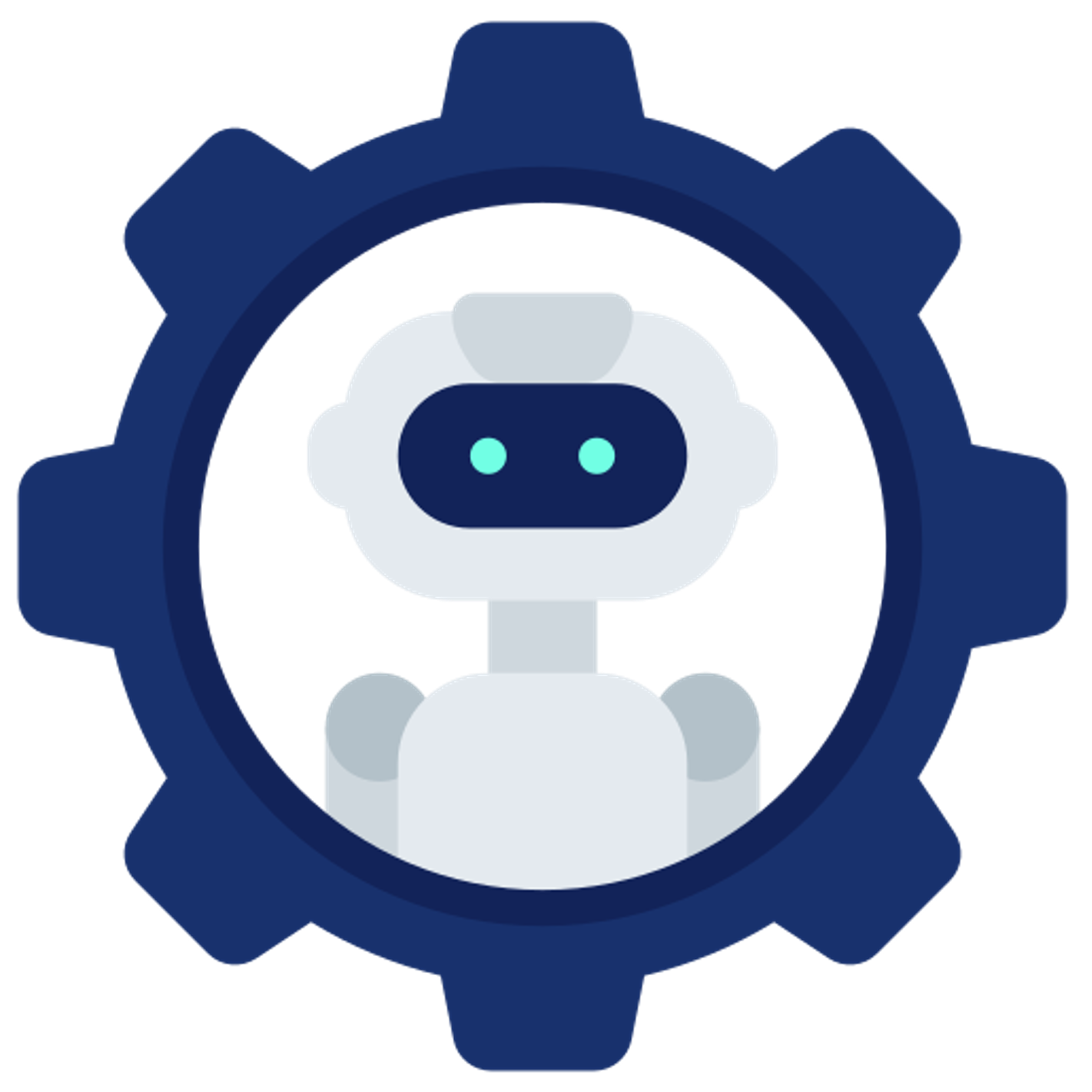} \textbf{Reflection \& Feedback (RF)}
        & \textbf{Drive Long-Term Learning}
        & Functions as the organizational learning facilitator. It compares the final prescription $\mathbf{P}_\textnormal{final}$ to a gold standard $\mathbf{P}_\textnormal{gold}$ to generate the discrepancy signal that drives the long-term adaptation of the entire system via the \textbf{DSWA} outer loop. \\
        [1ex]
        \hline\hline 
    \end{tabular}
   
    \label{tab:agent_roles}
\end{table*}

\section{Preliminary: Problem Definition}
Our work addresses the challenge of medication recommendation for multimorbid patients at the point of hospital admission. The core task is to generate a safe and effective prescription plan based solely on the clinical narrative available in the patient's record. This narrative, which we denote as input $\textbf{\textit{T}}$, comprises the main complaints of the patient $\textbf{\textit{CP}}$, personal and historical information $\textbf{\textit{PI}}$, and other risk factors $\textbf{\textit{OI}}$, such as allergies or lifestyle habits. The general input can be summarized as \textit{\textbf{T}} = \{ \textit{\textbf{CP}}, \textit{\textbf{PI}}, \textit{\textbf{OI}} \}.

The desired output is a complete prescription, $\textit{\textbf{P}}$, which includes both a list of drugs as well as a clinically sound interpretation. Specifically, a prescription consists of a drug list $\textit{\textbf{D}}$ and an accompanying interpretation $\textit{\textbf{I }}$. Each recommended drug $d_i$ in the list is detailed with the precautions $r_{i}$, the dosage and frequency $u_{i}$, and a clear explanation $e_i$ for its inclusion. Thus, the final output is \textit{\textbf{P}} = \{ \textit{\textbf{D}}, \textit{\textbf{I}} \}, where $\textbf{\textit{D}} = \{ (d_i, u_i, r_i, e_i) \}^{N}_{i=1}$.

\subsection{A Multi-Objective Optimization Problem}
As established in the Introduction, prescribing for multimorbid patients is fundamentally a coordination problem arising from fragmented clinical expertise. Here, we formalize the desired outcome of this coordination process. A successful prescription can be viewed as a solution to a multi-objective optimization problem, where the system must simultaneously satisfy several critical constraints related to safety, rationality, and interpretability.  The overall goal is to find an optimal prescription \textbf{\textit{P*}} that maximizes a set of objectives: \textbf{\textit{P*}} = $\arg\max_{\textit{P}}$ (Safety (\textit{P}), Rationality (\textit{P}), Interpretability (\textit{P})).

\textit{\textbf{Safety}} is the foremost concern. A valid prescription must be free of harmful \textbf{drug-drug interactions (DDI)} between any two medications in the plan. Furthermore, it must avoid \textbf{drug-patient interactions (DPI)}, ensuring that the medication does not conflict with the existing conditions or allergies of the patient. Finally, the \textbf{dosage} for every drug must be within a clinically accepted rational range.

\textit{\textbf{Pharmacological Rationality}} ensures that the prescription is both effective and efficient. This involves two main aspects. First, the plan must have \textbf{high coverage}, which means that all primary symptoms and therapeutic needs are addressed. Second, it must \textbf{avoid duplication}, where multiple drugs are prescribed for the same condition, leading to redundancy and increased risk. The goal is to maximize therapeutic coverage while simultaneously penalizing redundancy. This trade-off is perfectly captured by: \textbf{Rationality(P)} = \textbf{Coverage(P)} - \textbf{Duplication(P)}.

\footnotetext[1] {Formally, the safety checks correspond to satisfying the constraints: $\forall\, i \neq j: \textbf{DDI}_P(d_i, d_j) = 0$; $\forall\, i: \textbf{DPI}_P(d_i, \textbf{PI}_{n}) = 0$; and ensuring dosage $u_i \in [u^{min}_i,u^{max}_i]$.}
\footnotetext[2] {The rationality objective is formalized as maximizing coverage, where ideally $\forall cp \in \textbf{CP}$, $\exists d_i \in \textbf{D}$ s.t. $d_i \mapsto cp$ while ensuring zero duplication, $\forall i \neq j$, $Indication(d_i) \neq Indication(d_j)$ for the same complaint.}

\textbf{\textit{Interpretability}} requires that the final drug list should be transparent, traceable, and auditable by human experts. This is especially vital to ensure that nuanced patient-specific risk factors, which we categorize as Other Information (OI), have been properly considered. These factors include the patient's lifestyle (e.g., alcohol use), administration constraints (e.g., inability to swallow pills), or in a specific population with unique contraindications (e.g., pregnancy). Therefore, a complete solution must generate not just a prescription, but also clear evidence-based justifications for its recommendations.

\subsection{Baseline Approaches}
To benchmark our method, we compare it against two primary baseline paradigms that represent the current state-of-the-art in this domain:

\textbf{General-Purpose LLM Generation:} This approach uses a powerful, general-purpose large language model (e.g., GPT-4). The entire patient record \textbf{\textit{T}} is formatted into a prompt with specific strategies (zero-shot, few-shot, etc.), and the LLM is directly asked to generate a prescription \textbf{\textit{P}}. This method tests the out-of-the-box reasoning capabilities of a leading foundation model.

\textbf{Specialized Medical LLM Generation:} This paradigm utilizes a large language model that has been specifically developed for the medical domain. These models are pre-trained on extensive corpora of biomedical literature, clinical guidelines, and electronic health records. As a result, they possess deep domain knowledge and are expected to generate more clinically sound outputs than the general-purpose counterparts. The process remains end-to-end, taking \textbf{\textit{T}} as input to produce \textbf{\textit{P}}. This baseline represents a strong, knowledge-intensive competitor that leverages a state-of-the-art domain-expert model.

\section{Methodology -- \texttt{PRISMATIC}}
\begin{figure*}[htbp]
    \centering
    \includegraphics[width=0.8\linewidth]{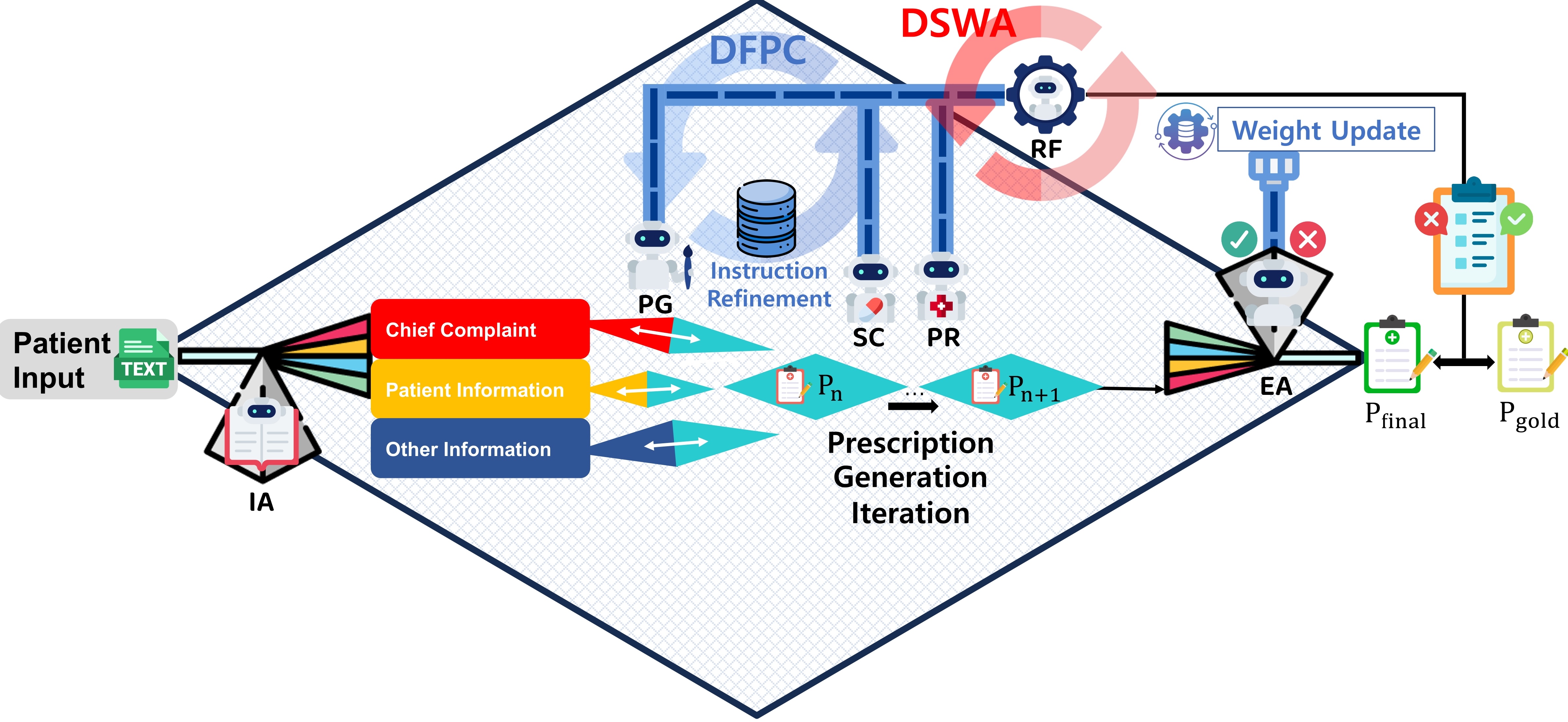}
    \caption{Complete workflow of \textbf{\texttt{PRISMATIC}}}
    \label{fig: WorkFlow}
    \Description{workflow}
\end{figure*}

To solve the multi-objective coordination problem of multimorbid prescribing, we introduce \textbf{\texttt{PRISMATIC}}, a multi-agent architecture. Conceptually, the architecture acts as an intelligent prism: it takes unstructured clinical text as input, "refracts" it through the specialized perspectives of multiple agents, and synthesizes their insights into a single, coherent prescription. The whole process is governed by a dual-loop learning design:
\begin{enumerate}
    \item An \textbf{Inner Loop (DFPC)} handles the collaborative reasoning and refinement required to generate a prescription for a single patient case.
    \item An \textbf{Outer Loop (DSWA)} enables long-term, system-wide adaptation and learning across multiple cases.
\end{enumerate}
The architecture consists of six specialized agents whose roles are explicitly designed to address the objectives defined in our problem formulation. The conceptual architecture is shown in Figure \ref{fig: WorkFlow}.

\subsection{The Agents: A Society of Clinical Specialists}
The \texttt{PRISMATIC} agent society simulates a multidisciplinary clinical team. Each agent has a distinct responsibility, and their collaboration is mediated through a shared blackboard. Their roles are summarized in Table \ref{tab:agent_roles}.
\subsection{The Inner Loop: Case-Level Refinement (DFPC)}
The Discrepancy Feedback Precise Calibration (DFPC) mechanism governs the real-time collaboration for a single patient case. It is an iterative \textbf{propose-critique-refine} cycle:

\textbf{\textit{Propose:}} The \textbf{PG} agent initiates a candidate prescription.

\textbf{\textit{Critique:}} The \textbf{SC} and \textbf{PR} agents evaluate it concurrently against the safety and rationality objectives, posting their findings.

\textbf{\textit{Adjudicate \& Refine:}} The \textbf{EA} agent aggregates this feedback.If the prescription is unacceptable, it triggers a refinement cycle.

The key to DFPC's efficiency and effectiveness lies in its mechanism for targeted error attribution and local calibration. Instead of restarting the generation process, the system traces a specific violation (e.g., a \textbf{DDI}) back to its source domain and re-engages only the relevant agents (e.g., the \textbf{SC} and \textbf{PG}). The \textbf{SC} provides specific feedback, and the \textbf{PG} performs a localized correction—such as replacing a single conflicting drug—while leaving the rest of the prescription intact. This process of decentralized refinement makes the multi-agent deliberation both computationally efficient and clinically stable.

\subsection{The Outer Loop: Long-Term Adaptation (DSWA)}
The Dynamic Self-updating Weight Adjustment (DSWA) mechanism enables the system to learn and improve over many cases. After the inner loop finalizes a prescription, the \textbf{RF} agent identifies discrepancies between the output ($\mathbf{P}_\textnormal{final}$) and a gold-standard reference ($\mathbf{P}_\textnormal{gold}$). This error signal drives the DSWA process in two ways:

First, it performs \textbf{adaptive weight evolution}. DSWA quantifies the errors across different clinical dimensions (e.g., DDI safety, dosage, coverage) and updates a global weight vector, \textbf{W}, used by the \textbf{EA} agent. Clinical dimensions that exhibit persistent or severe errors are systematically up-weighted, which compels the agent society to prioritize those aspects more heavily in future deliberations.

Second, \textbf{DSWA} triggers a \textbf{distributed knowledge consolidation}. Insights from the error analysis are distilled and synchronized across the agents' knowledge bases, allowing the entire system to learn from cases. This dual process of adapting the collective evaluation policy (\textbf{W}) and updating shared knowledge enables the system to develop an emergent understanding of clinical priorities, mirroring how a human team's expertise evolves with experience.

This process functions as a form of organizational learning, allowing the agent society to adapt its collective priorities over time and become more attuned to the specific challenges of its clinical environment.

\begin{algorithm}
\caption{Dynamic Self-updating Weight Adjustment (DSWA)}

\label{alg:dswa}
\begin{flushleft}
\textbf{Input:} Discrepancy vector $\mathbf{r}^{(t)} = [r_1^{(t)}, \dots, r_6^{(t)}]$; previous weights $\mathbf{W}^{(t)} = [W_1^{(t)}, \dots, W_6^{(t)}]$; smoothing factor $\gamma$; learning rate $\eta$; temperature $\beta$.\\
\textbf{Output:} Updated weights $\mathbf{W}^{(t+1)} = [W_1^{(t+1)}, \dots, W_6^{(t+1)}]$.\\[4pt]

\textbf{Temporal smoothing:}\\[-4pt]
\begin{equation*}
\bar{r}_i^{(t)} \leftarrow (1 - \gamma)\bar{r}_i^{(t-1)} + \gamma r_i^{(t)}
\end{equation*}

\textbf{Compute normalized importance:}\\[-4pt]
\begin{equation*}
u_i^{(t)} \leftarrow 
\frac{\exp(\beta\,\bar{r}_i^{(t)})}
{\sum_j \exp(\beta\,\bar{r}_j^{(t)})}
\end{equation*}

\textbf{Update weights:}\\[-4pt]
\begin{equation*}
\tilde{W}_i^{(t+1)} \leftarrow (1-\eta)W_i^{(t)} + \eta u_i^{(t)}
\end{equation*}

Normalize:\\[-4pt]
\begin{equation*}
W_i^{(t+1)} \leftarrow \frac{\tilde{W}_i^{(t+1)}}{\sum_j \tilde{W}_j^{(t+1)}}
\end{equation*}

\textbf{Return:} Updated $\mathbf{W}^{(t+1)}$
\end{flushleft}
\end{algorithm}

\begin{table*}[!t]
  \centering
  \caption{Comparison of performance of mainstream large models in medical question-answering tasks under different complexity levels. Complexity $i$ to $iii$ represents simple to complicated tasks. This table summarizes the performance of four models under various strategies and SFT medical models. \textbf{Bold values} indicate the best performance for each complexity level. \underline{Underline values} indicates the second-best performance for each complexity.}
  \fontsize{9pt}{11pt}\selectfont
  \setlength{\tabcolsep}{3pt}
  \renewcommand{\arraystretch}{0.5}
  \begin{tabular}{cc|cccc|cccc|cccc|cccc}
    \toprule
    & & \multicolumn{4}{c}{\textbf{Complexity \romannumeral1(\%)}} 
    & \multicolumn{4}{c}{\textbf{Complexity \romannumeral2(\%)}} 
    & \multicolumn{4}{c}{\textbf{Complexity \romannumeral3(\%)}}
    & \multicolumn{4}{c}{\textbf{Complete Dataset(\%)}}\\
    \cmidrule(lr){3-6} \cmidrule(lr){7-10} \cmidrule(lr){11-14} \cmidrule(lr){15-18}
    
    \textbf{Method} & \textbf{Model} 
    & OR & Prec. & EM  & ZIR 
    & OR & Prec. & EM  & ZIR 
    & OR & Prec. & EM  & ZIR
    & OR & Prec. & EM  & ZIR\\

    \midrule
     
    \multirow{4}{*}{Zero-Shot} 
    & GPT-4o        &45.87 &51.89 & 4.08   &95.24  &35.20  &36.89  &2.37    &93.01  &15.35  &19.35  &0.34  &81.47 &29.11&25.72&2.05&89.13 \\
    & Qwen-3       &41.12  &44.46 &4.01    &95.31  &32.45  &38.90  & 0.99  &92.40  &16.09  &20.12  &0.43  &81.80  &21.56&28.78&1.60&89.08 \\
    & DeepSeek-R1  &47.23  &48.06  &10.20 &94.43  &39.13  &37.02  & 1.27   &93.06  &17.89 &14.01  &0.43 &78.74  &31.48&25.20&3.39&87.87 \\
    & LLaMA3.1    &45.10 &43.29  &6.05 &94.49  &29.11  &31.78  & 1.60   &91.19  &13.90 &19.91  &0.10 &79.74 &32.45 & 36.40 &2.23&87.65 \\
    \cmidrule(l){2-18}

    \multirow{4}{*}{Few-Shot} 
    & GPT-4o &44.23  &58.91  &7.27 &95.92 &29.09 &31.98  &1.38 &93.06  &19.89  &20.11  &0.29    &86.78 &31.61&29.09&2.57&91.40  \\
    & Qwen-3 &41.22  &45.90  &8.09 &95.92 &28.91  &27.56 &1.54    &94.27  &17.89 &19.10  &0.43 &84.05 &32.14&35.09&2.90&90.75 \\
    & DeepSeek-R1  &48.10  &47.09  &8.84 &94.70  &32.10  &38.12  &2.15    &94.16  &20.19  &22.75 &0.38    &86.64  &32.24&40.11&3.29&91.39\\
    & LLaMA3.1  &44.23  &43.09 &12.10 & 95.38 &29.78 &25.45  &1.49    &92.24  &16.78 &17.15 &0.38    &78.30 &26.61&32.09&3.96&87.68 \\
    \cmidrule(l){2-18}

    \multirow{4}{*}{CoT} 
    & GPT-4o &49.23  &53.17  &11.76 &96.06 &34.56 &39.09  &1.65    &94.16  &16.70  &18.94  &0.34    &89.32 &37.40&44.30&3.91&92.80  \\
    & Qwen-3       &46.76  &47.88  & 13.66   &95.31  &28.78  &27.90  & 2.70   &95.48  &17.55  &21.11  &0.62    &79.74  &35.81& 39.88 &4.89&89.32 \\
    & DeepSeek-R1  &51.40  &49.20  &12.85    & 94.43 &34.56  &34.23  &1.82   &97.80 &19.10 &21.88  &0.43    &79.31  &31.09&32.23&4.30&89.69 \\
    & LLaMA3.1    &42.19  &48.55  &13.05    & 95.45 &20.19  & 29.09 &2.09    &95.32 &22.31  &18.71  &0.19    &81.32  &36.31&28.61&4.35&89.92 \\
    \cmidrule(l){2-18}

    \multirow{4}{*}{ToT} 
    & GPT-4o        &59.12  &67.23  &23.18    &97.21  &39.01  &38.76  &2.26    &97.25  &25.22 &24.89  &0.77 &85.34 &31.01&33.88&7.40&92.61  \\
    & Qwen-3       &54.23  &53.03  &22.71    &96.67  &37.87  &42.88  &1.16   &96.70  &24.09 &21.09  &0.91   &81.42  &32.34&29.81&6.90&90.75 \\
    & DeepSeek-R1  &58.92  &59.03  &23.45    & 98.91 &44.87  &39.10  &1.05    &98.40  &23.09  &28.91  &0.81    &85.58 &35.89&34.11&7.09&93.56  \\
    & LLaMA3.1    &60.23  &47.12  &17.27    & 97.42 &35.89  &29.12  &1.87    &96.70  &19.09 &24.35 &0.62    &86.78 &28.34&29.01&5.60&93.04 \\
    \cmidrule(l){2-18}

    \multirow{4}{*}{ReAcT} 
    & GPT-4o        &53.09  & 59.13 &22.71   &95.92  &39.13  &41.90  &1.76    &94.27  &31.89  &28.90  &0.67 &85.68  &37.09&34.89&7.07&91.39 \\
    & Qwen-3       &54.09  &57.34  &21.75    & 95.38 &37.71  &35.64  &1.60   &93.01  &32.90  &35.13  &0.34    &82.28   &38.90&38.10&6.62&89.49\\
    & DeepSeek-R1  &53.17  &52.94  &21.10    &96.53  &33.45  &39.12  &2.59    &92.62  &31.87  &40.19  &0.53    &86.59  &40.89&45.34&6.83&91.35\\
    & LLaMA3.1    &51.98  &54.07  &20.33    &95.04  &31.11  &39.77  & 1.98   &92.02  & 30.21 &33.01  &0.29   &84.82  &39.98&41.28&6.34&90.05 \\
    \cmidrule(l){2-18}
    
    \multirow{4}{*}{MedPrompt} 
    & GPT-4o        &61.11  &70.30  &20.46 &98.98  &53.98  &59.09  &4.90    & 94.49 &48.77  &43.91  &1.44    &92.42 &51.30&49.11&7.81&94.92 \\
    & Qwen-3       &59.31  &71.24  &22.71 &99.25 &51.82  &59.32  &3.69    &93.67  &40.34  &44.71  &0.43   &83.91 &45.01&51.03&7.63&91.40  \\
    & DeepSeek-R1  &67.34  &75.34  &23.18    &\underline{99.46}  &45.99  &54.34 &\underline{4.52}  &96.63  &43.98  &49.01  &0.38   &80.60 &49.30&52.34&8.02&91.18  \\
    & LLaMA3.1    &67.30  &69.31  &21.62    &99.18  &58.90  &53.78  & 3.36   &94.71  &39.10  &43.10  &0.57   &87.74 &45.19&50.01&7.27&89.53 \\
    \cmidrule(l){2-18}

    \multirow{4}{*}{OpenMedLM} 
    & GPT-4o        &60.12  &69.23  &24.95 &98.16  &43.90 &40.01  &2.26 &94.77  &38.11  &35.01  &\underline{1.53}  &92.24 &39.90&40.57&8.19&94.72  \\
    & Qwen-3       &57.34  &70.30  &24.81    &98.98  &45.90  &43.01  &1.71    &94.16  &31.23  &35.12  &0.24    & 83.09 &39.13&38.10&7.46&91.18 \\
    & DeepSeek-R1  &68.13 &71.17  &25.22 &99.39  &42.98  &38.78  &2.26 &96.04  &33.78  &32.09  &0.43   & 95.40  &43.13&40.94&7.83&96.71 \\
    & LLaMA3.1   &65.10  &65.30  &25.76 &97.96  &39.01  &35.76  &0.22    &96.04  &31.09  &32.88  &0.57   &77.92 &40.88&41.97&7.27&89.53 \\
    \midrule

    \multirow{3}{*}{SFT} 
    & Meditron &72.01 &74.47 &26.51 & 98.57 &53.89 &53.81  &4.30 &96.96  &31.80  &38.01  &0.48    &92.10 &46.09&42.11&8.89&95.48   \\
    & PMC &73.32  &72.91  & 32.63   &97.48  &49.67  &48.91  &2.26    &97.19  &38.78  &34.09  &0.43   &88.07  &47.55&48.10&9.86&93.73 \\
    & Med42 &\textbf{78.12}  &\underline{84.69}  &31.48  &98.98 &55.87  &53.40  &1.71    &\underline{98.40}  &\textbf{49.92}  &44.11  &0.38    &89.75 &51.20&53.98&9.34&95.20   \\
    \midrule
    
    \multirow{4}{*}{\textbf{\texttt{PRISMATIC}}} 
    & GPT-4o &77.22  &81.22  &\textbf{42.22} &99.25 &51.09  &\underline{62.32}  &4.02    &97.85  &41.23  &43.98  &\textbf{1.72}    &\underline{97.75}  &\textbf{56.81}&\underline{60.11}&\textbf{13.58}& \underline{98.20}\\
    & Qwen-3 &75.09  &80.32  &29.16    &99.18  &\underline{59.78} &57.61  &4.30    &97.19  &41.34  &49.78  &0.96    & 96.12 &49.13&55.81&9.80&97.32 \\
    & DeepSeek-R1 &\underline{77.92}  &\textbf{84.90}  &\underline{37.59}    &\textbf{99.66}  &\textbf{61.98}  &\textbf{68.11}  &4.07    &\textbf{98.51}  &\underline{48.81}  &\textbf{52.77}  &0.86    & \textbf{99.47} &\underline{56.01}&\textbf{62.23}&\underline{12.00}&\textbf{99.20} \\
    & LLaMA3.1 &71.01 &76.09  &30.66    &99.12  &52.13 &62.03    &\textbf{5.29}    &97.96  &41.44  &\underline{49.80}  &0.67    &96.55 &51.40&56.70&10.44&97.73 \\

    \bottomrule
  \end{tabular}
  \label{tab:model_performance}
\end{table*}

\section{Experiments}
\subsection{Experiment Setup}
\textbf{Dataset and Task.} We evaluate PRISMATIC on a curated dataset from MIMIC-IV, consisting of pairs of unstructured admission notes (\textbf{\textit{T}}) and their corresponding ground-truth prescriptions ($\mathbf{P}_\textnormal{gold}$). To reflect real-world clinical challenges, we stratified the dataset into three levels of case complexity (low, medium, high) based on the number of chronic diagnoses and prescribed medications per patient with K-means, which were denoted as complexity \romannumeral1, \romannumeral2, and \romannumeral3, respectively.

\textbf{Evaluation Metrics.} We assess performance from two critical perspectives:
\begin{enumerate}
    \item \textbf{Accuracy and Rationality:} We use Overlap Rate (OR), Precision (Prec.), and Exact Match (EM) to measure how closely the generated prescriptions match the ground truth. EM is the strictest metric, requiring a perfect match of all prescribed drugs.
    \item \textbf{Safety:} We use the Zero-Interaction Ratio (ZIR), which measures the percentage of generated prescriptions that are completely free of known drug-drug interactions. This metric directly evaluates the system's ability to satisfy a primary clinical objective.
\end{enumerate}
\begin{itemize}
    \item \textbf{Overlap Rate (OR).}  
    Measures the proportion of gold-standard drugs correctly predicted. For the $i$-th case:
    \[
    \text{OR} = \frac{1}{n} \sum_{i=1}^{n}
    \frac{|\mathcal{\textbf{P}}_i \cap \mathcal{\textbf{P}}_{gold}^{(i)}|}{|\mathcal{\textbf{P}}_{gold}^{(i)}|}
    \]

    \item \textbf{Precision (Prec.)}  
    Measures the proportion of predicted drugs that are correct:
    \[
    \text{Prec.} = \frac{1}{n} \sum_{i=1}^{n}
    \frac{|\mathcal{\textbf{P}}_i \cap \mathcal{\textbf{P}}_{gold}^{(i)}|}{|\mathcal{\textbf{P}}_i|}
    \]

    \item \textbf{Exact Match Ratio (EM).}  
    Measures the fraction of predictions that exactly match the ground-truth set:
    \[
    \text{EM} = \frac{1}{n} \sum_{i=1}^{n} \mathbb{I}[\mathcal{\textbf{P}}_i = \mathcal{\textbf{P}}_{gold}^{(i)}]
    \]
    where $\mathbb{I}[\cdot]$ is the indicator function returning 1 if the predicted and gold-standard sets are identical, and 0 otherwise.
    \item \textbf{Zero-Interaction Ratio (ZIR).}  
    Measures the fraction of prescriptions that are entirely free of drug-drug interactions:
    \[
    \text{ZIR} = \frac{1}{n} \sum_{i=1}^{n} 
    \mathbb{I}\left[
    [\forall (d_a, d_b)] \in \binom{\mathcal{\textbf{P}}_i}{2},\notin \mathcal{\textbf{D}}
    \right]
    \]
    where $\mathbb{I}[\cdot]$ is the indicator function returning 1 if no interacting drug pair is found in $\mathcal{\textbf{P}}_i$, and 0 otherwise.
\end{itemize}

\textbf{Baseline Methods.} To demonstrate the superiority of our multi-agent architecture, we compare \texttt{PRISMATIC} against a comprehensive suite of strong baselines representing two main paradigms:

\begin{enumerate}
    \item \textbf{General-Purpose LLMs:} We test the out-of-the-box reasoning of state-of-the-art models (GPT-4o, Llama-3.1, Qwen3, DeepSeek-R1) using a range of prompting strategies, including Zero-Shot, Few-Shot \citep{brown2020languagemodelsfewshotlearners}, Chain-of-Thought \citep{wei2023chainofthoughtpromptingelicitsreasoning}, Tree-of-Thoughts \citep{yao2023treethoughtsdeliberateproblem}, and ReAct \citep{yao2023reactsynergizingreasoningacting}. Besides, medical prompt methods, including MedPrompt \citep{chen2023medpromptcrossmodalpromptingmultitask} and OpenMedLM \citep{maharjan2024openmedlmpromptengineeringoutperform} Prompt, are also considered.
    \item \textbf{Specialized Medical Models:} We include models specifically designed for the medical domain, including Meditron \citep{chen2023meditron70bscalingmedicalpretraining}, PMC-LLaMA \citep{wu2023pmcllamabuildingopensourcelanguage}, and LLaMA-3.1-Med42 \citep{christophe2024med42v2suiteclinicalllms}, which represent monolithic and knowledge-intensive competitors.
\end{enumerate}

\textbf{\texttt{PRISMATIC} Implementation.} To ensure a fair and direct comparison, the agents within the \texttt{PRISMATIC} architecture were implemented using the same suite of base models as the baselines. For example, the row labeled PRISMATIC (GPT-4o) in our results table (Table 2) signifies an instance of our system where all agents (e.g., PG, SC, PR) utilize GPT-4o for their reasoning and generation tasks. This experimental design allows us to isolate the performance gains attributable to our multi-agent coordination and learning architecture, directly contrasting its efficacy against monolithic prompting of the exact same underlying LLM.

\subsection{Main Results}
\texttt{PRISMATIC} consistently and significantly outperforms all baseline models across all metrics, with its advantage being most pronounced in safety and in managing high-complexity cases, as shown in Table~\ref{tab:model_performance}.

\textbf{Dominance in Clinical Safety.} \texttt{PRISMATIC} achieves a new standard of safety in the generation of prescriptions. In the complete data set, it reaches a zero interaction ratio (ZIR) of 99. 20\%, virtually eliminating drug-drug interactions. This stands in stark contrast to the baselines; even the strongest specialized models like Meditron struggle to surpass 95\%, while many advanced prompting methods fall below 93\%. This dramatic improvement is a direct validation of the Safety Check (SC) agent, which functions as an explicit safety regulator, a mechanism fundamentally absent in the monolithic design of baseline models.

\begin{figure}[h]
    \centering
    \includegraphics[width=1\linewidth]{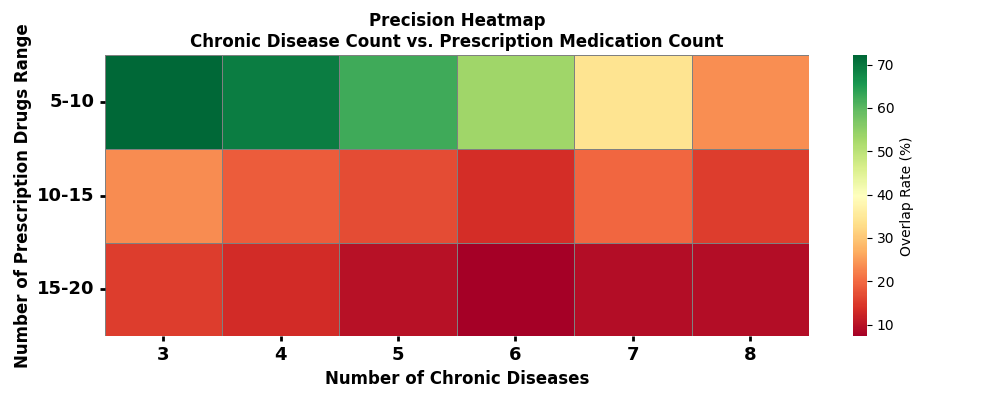}
    \caption{Heatmap of precision across varying chronic disease and prescription complexities.}
    \label{fig:orhp}
    \Description{heatmap complexity}
\end{figure}

\textbf{Superior Accuracy in High-Complexity Cases.} \texttt{PRISMATIC}'s superiority extends to accuracy and pharmacological rationality. The escalating difficulty of this task is visualized in the performance heatmap in Figure \ref{fig:orhp}. As the number of chronic diseases and prescribed medications increases, corresponding to the green-to-red transition in the heatmap, all models are challenged. However, while the monolithic baselines suffer a catastrophic performance decay in these high-complexity scenarios, \texttt{PRISMATIC}'s performance degrades far more gracefully. It maintains high accuracy and rationality (OR, Prec., and EM). Its 13.58\% EM score on the complete dataset, more than double the next best baseline, highlights its robust ability to handle the combinatorial challenges of multimorbid prescribing. This supports our central hypothesis: a multi-agent coordination approach is fundamentally better suited for complex, multi-faceted problems than a monolithic one.

\subsection{Ablation Study}

To validate our architectural design and quantify the contribution of each component, we conducted a comprehensive ablation study using GPT-4o as the base model. The results, presented in Table \ref{tab:ablation}, confirm that \texttt{PRISMATIC}'s success is an emergent property of its multi-agent synergy, rather than a function of any single part.

\textbf{The Critical Role of Coordination and Learning Mechanisms.} The study reveals that the dual-loop learning mechanisms are the most critical drivers of overall prescription accuracy. Removing the inner-loop refinement protocol, \textbf{DFPC}, causes the most significant drop in overall performance, reducing the exact match (EM) score by nearly half, from 13.58\% to 7.42\%. This highlights that iterative, multi-agent negotiation is the single most important factor in achieving a complete and correct prescription. Similarly, removing the long-term adaptation mechanism, \textbf{DSWA}, also severely degrades performance, with the EM score falling to 8.56\%. This confirms the necessity of organizational learning for mastering the complexities of the task.

\textbf{Indispensable Contributions of Specialist Agents.} The ablations of the specialized "critic" agents provide irrefutable evidence for their roles:
\begin{enumerate}
    \item Removing the \textbf{Safety Check (SC)} agent is catastrophic for patient safety. The Zero Interaction Ratio (ZIR) falls from a near-perfect 98. 20\% to an unacceptable 78.98\%. This 19-point drop is a clear and powerful demonstration that an explicit, dedicated safety agent is non-negotiable for generating clinically safe prescriptions.
    \item Removing the \textbf{Pharmacological Rationality (PR)} agent significantly harms the prescription's therapeutic value, causing the \textbf{Overlap Rate (OR)} to fall from 56.81\% to 44.13\%. This validates its essential function in ensuring that prescriptions are both comprehensive and efficient.
\end{enumerate}

The ablation study demonstrates that each agent and mechanism in the system contributes uniquely within its designated scope, collectively enhancing the overall performance of the prescription generation framework. Overall, each component fulfills its role and adapts seamlessly to the multi-agent workflow, resulting in a robust, safe, and clinically rational prescription generation system.

\begin{table}[t]
\centering
\caption{Ablation study results for the \textbf{\texttt{PRISMATIC}} model with GPT-4o across various components. \textbf{Bold values} indicate the validity of \textbf{\texttt{PRISMATIC}}.}
\fontsize{9pt}{11pt}\selectfont
\begin{tabular}{ccccc}
\toprule
\textbf{Component}  & \textbf{OR} & \textbf{Prec.} & \textbf{EM}  & \textbf{ZIR} \\
\midrule
w/o ~\includegraphics[height=1em]{Agents/IA.png} \textbf{IA}  &52.43  &59.34 &11.98   &93.43  \\
w/o ~\includegraphics[height=1em]{Agents/SC.png} \textbf{SC}   &41.25  &50.22  &10.56  &78.98 \\
w/o ~\includegraphics[height=1em]{Agents/PG.png} \textbf{PR}   &44.13  &49.65  &9.33  &82.91  \\
\midrule
w/o ~\includegraphics[height=1em]{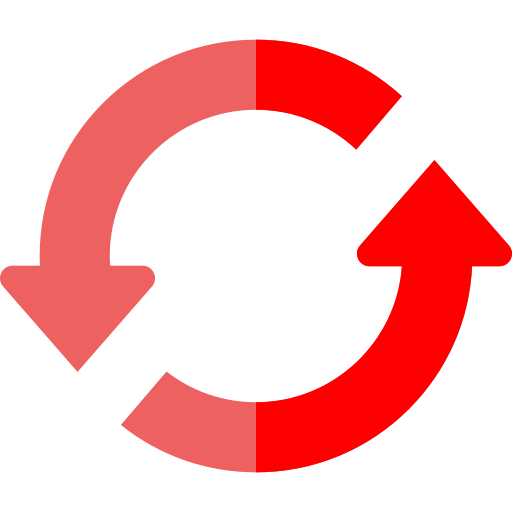} \textbf{DSWA}  &48.55 &55.11  &8.56 &91.23 \\
\midrule
w/o ~\includegraphics[height=1em]{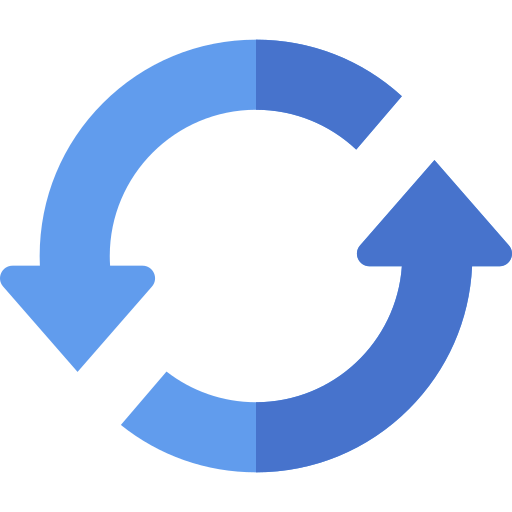} \textbf{DFPC}  &49.56   &59.22  &7.42 &89.35 \\
\midrule
w/o ~\includegraphics[height=1em]{Agents/EA.png} \textbf{EA}  &54.90  &67.23  &12.78 &95.09\\
w/o ~\includegraphics[height=1em]{Agents/RF.png} \textbf{RF}   &55.89  &58.06 &12.87 &96.00 \\
\midrule
~\includegraphics[height=1em]{prism.png}\textbf{\texttt{PRISMATIC}}  &\textbf{56.81}  &\textbf{60.11}  &\textbf{13.58} &\textbf{98.20} \\
\bottomrule
\end{tabular}

\label{tab:ablation}
\end{table}

\section{Analysis and Discussion}
\subsection{Emergent Learning and Priority Adaptation in DSWA}
\begin{figure}[htbp]
    \centering
    \includegraphics[width=1\linewidth]{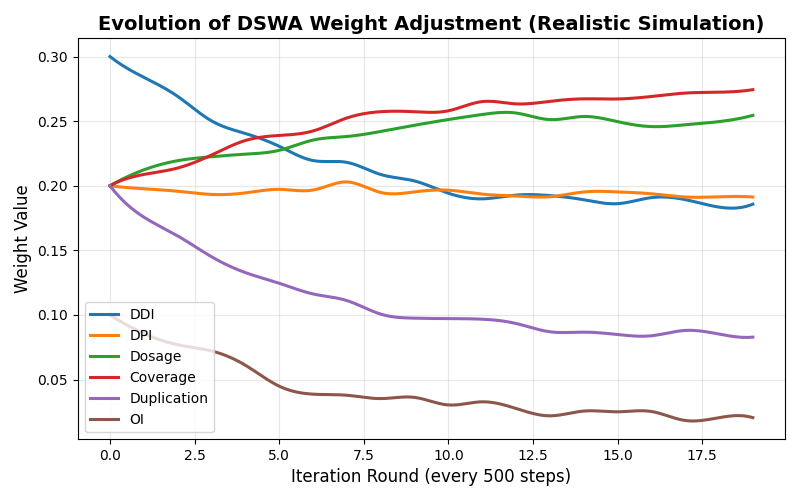}
    \caption{ Evolution of risk weights across six dimensions in the DSWA mechanism. Every 500 cases per iteration round. The line graph illustrates how weights dynamically adjust over time.}
    \label{fig:weight}
    \Description{DSWA weight change.}
\end{figure}

A core claim of our work is that \texttt{PRISMATIC} performs organizational learning. Figure \ref{fig:weight} provides direct visual evidence of this emergent behavior by plotting the evolution of the six objective weights governed by the DSWA mechanism.

The graph visualizes the clear and intelligent adaptation process of the system. Initially, all weights start from a preset value, which is adopted from . However, over thousands of cases, the system learns to prioritize different clinical objectives based on historical error patterns. The weights for Dosage and Coverage steadily increase, indicating that these are complex, nuanced aspects where the system frequently makes initial errors that require correction. In response, the DSWA mechanism adaptively up-weights their importance, compelling the agent society to allocate more attention to them. Conversely, the weight for Duplication gradually decreases, suggesting this is a more straightforward constraint that the system learns to satisfy relatively quickly.

Crucially, the safety-related weights (DDI and DPI) remain consistently high, reflecting their non-negotiable importance. This dynamic reallocation of priorities is a hallmark of adaptive learning. The system autonomously identifies its own weaknesses and strategically adjusts its collective evaluation policy, demonstrating an emergent intelligence that monolithic, static models lack.

\subsection{Error Analysis: The Challenge of Replicating Expert Judgement}
Although \texttt{PRISMATIC} significantly outperforms all baselines, its absolute Exact Match (EM) score of 13.58\% underscores the profound difficulty of fully replicating the nuanced decision-making of human clinicians. We denoted this gap between the system's output and the ground truth stems from several sophisticated clinical challenges:
\begin{enumerate}
    \item \textbf{Omission of Ancillary Medications:} The system excels at identifying core treatments but sometimes omits secondary medications for symptomatic relief or preemptive care (e.g., anti-nausea drugs). These adjunct medications, while not central, are crucial for holistic patient care.
    \item \textbf{Suboptimal Selection Among Equivalent Drugs:} When faced with multiple drugs with similar pharmacological effects, the system may not select the specific drug a clinician would prefer based on implicit factors such as patient tolerance, cost, or local hospital formulary.
    \item \textbf{Difficulty with Implicit Clinical Context:} The system is limited to the explicit text of the clinical note. It cannot yet infer the subtle, unwritten context or draw upon the years of experiential knowledge that inform an expert's "clinical intuition."
\end{enumerate}

These observations are illustrated with concrete examples in the Appendix, where specific cases highlight the circumstances under which each of these defects manifests.

These limitations are not merely model defects but represent the frontier of clinical AI. Addressing them will require future work on more advanced reasoning mechanisms, integration with diverse knowledge sources (e.g., pharmacoeconomics), and new agent designs capable of handling implicit and commonsense clinical knowledge.

\section{Conclusion}
This paper addressed the critical challenge of generating safe and rational prescriptions for multimorbid patients by reframing it as a multi-agent coordination problem. We introduced \texttt{PRISMATIC}, a novel architecture whose performance validates this perspective. By assigning specialized reasoning tasks to distinct agents and governing their interactions with two learning mechanisms: case-level refinement (\textbf{DFPC}) and long-term policy adaptation (\textbf{DSWA}), our system effectively manages the competing objectives inherent in clinical practice. \texttt{PRISMATIC} successfully navigates the trade-offs between clinical safety, efficacy, and rationality. Its decisive outperformance of strong baseline models, especially in eliminating safety risks and mastering high-complexity cases, provides compelling evidence for the superiority of this agent-based paradigm. \texttt{PRISMATIC} serves as a blueprint for building more robust and interpretable clinical AI, suggesting that structuring complex reasoning as a collaborative process is a promising path forward. It also indicates that the future of AI in high-stakes domains like medicine may lie not in ever-larger models, but in smarter multi-agent coordination.

\section{Acknowledgment}
The authors thank anonymous colleagues for their helpful discussions. This work received no external funding. \\
The authors used AI-assisted writing tools for language editing purposes only. All content and ideas were developed by the authors.

\section{Ethics}
This study does not involve direct experimentation on human or animal subjects. All data used were either publicly available or properly anonymized to ensure that no personally identifiable information (PII) was involved. 
This study uses the MIMIC-IV database, a publicly available, de-identified dataset of critical care patients. Access to MIMIC-IV is governed by the PhysioNet Credentialed Health Data License 1.5.0, which permits non-commercial research use under strict conditions to protect patient privacy. All authors complied with the data usage agreement and completed the required human subjects research training. No attempt was made to re-identify any individuals. The use of this dataset adheres to relevant ethical guidelines and institutional standards.




\bibliographystyle{ACM-Reference-Format} 
\bibliography{sample}


\end{document}


\appendix

\section{Dataset and Task Details}

\label{app:dataset}
\subsection{Dataset}
MIMIC-IV (Medical Information Mart for Intensive Care IV) is a large, publicly available electronic health record (EHR) dataset collected from the Beth Israel Deaconess Medical Center between 2008 and 2019. It contains de-identified health-related data associated with over 40,000 patients who stayed in the hospital’s critical care units. The dataset includes structured information such as demographics, vital signs, laboratory tests, procedures, medications, and diagnostic codes. MIMIC-IV is designed to support a wide range of research in clinical decision-making, patient modeling, and machine learning applications in healthcare.\\
MIMIC-IV is organized into modular components, including the mimic-core, mimic-hosp, and mimic-note modules. \textbf{The MIMIC-IV-hosp} module contains detailed hospital-level data, such as medication orders, laboratory measurements, microbiology results, imaging procedures, and pharmacy prescriptions. It enables researchers to study treatment patterns and medical interventions during hospital stays.\\
The \textbf{MIMIC-IV-note} module provides unstructured free-text clinical notes written by physicians, nurses, and other healthcare providers. These include admission and discharge summaries, progress notes, radiology reports, and other unstructured documentation. MIMIC-IV-note is especially valuable for natural language processing (NLP) tasks aimed at extracting clinical information and understanding the context of care.

\subsection{Data Details}
We evaluate \textbf{\texttt{PRISMATIC}} on the MIMIC-IV patient cases that built upon unstructured clinical texts from MIMIC-IV-note and structured hospital data from MIMIC-IV-hosp. Here are the task data setup details in Table~\ref{tab:task-setup}.

\begin{table}[h]
\centering
\small
\renewcommand{\arraystretch}{1.2}
\begin{tabular}{>{\centering\arraybackslash}p{2.2cm} | p{13.2cm}}
\toprule
\textbf{Category} & \textbf{Details} \\
\midrule

\textbf{Data Using} & 
\texttt{MIMIC-IV-note} (discharge): \texttt{subject\_id}, \texttt{hadm\_id}, \texttt{text} \\
& \texttt{MIMIC-IV-hosp} (admissions): \texttt{admittime} \\
& \texttt{MIMIC-IV-hosp} (prescriptions): \texttt{starttime} \\
& \texttt{MIMIC-IV-hosp} (admissions): \texttt{subject\_id}, \texttt{hadm\_id} \\
& \texttt{MIMIC-IV-hosp} (diagnoses\_icd): \texttt{subject\_id}, \texttt{hadm\_id}, \texttt{icd\_version}, \texttt{icd\_code} \\
& \texttt{MIMIC-IV-hosp} (hospital\_prescriptions): \texttt{subject\_id}, \texttt{hadm\_id}, \texttt{drug}, \texttt{gsn}, \texttt{ndc} \\

\midrule
\textbf{Structured Data Cleaning} & 
Step 1: Patients with prescription records in \texttt{hosp} that met our screening criteria and had clinically owned text in \texttt{notes} were screened. \\
& Step 2: Merge \texttt{icd\_code} by \texttt{subject\_id} and \texttt{hadm\_id}. \\
& Step 3: Select patients with 3--8 chronic diseases based on \texttt{icd\_code} and \texttt{icd\_version}. \\
& Step 4: Merge \texttt{drug}, \texttt{gsn}, and \texttt{ndc} into a medication list. \\
& Step 5: Create variables \texttt{numdrug} and \texttt{numchronic}. \\

\midrule
\textbf{Unstructured Data Extraction} &
Extracted from \texttt{MIMIC-IV-note} (if exists):\\
& \quad - Chief complaint \\
& \quad - Allergies \\
& \quad - Invasive Procedure \\
& \quad - History of Present Illness \\
& \quad - Past medical history \\
& \quad - Physical exam \\
& \quad - Family history \\
& \quad - Social history \\
& \quad - Medications on Admission \\
& \quad - Discharge Diagnosis \\

\bottomrule
\end{tabular}
\caption{Task Setup Details and Data Extraction}
\label{tab:task-setup}
\end{table}

We'd like to clarify that we do not use full discharge summaries as model input. While the MIMIC-IV-NOTE corpus includes various types of unstructured notes, we only extract a single field—the “discharge diagnosis” section—from the discharge note, and we treat this as a proxy for the clinically established primary diagnosis. This diagnosis typically becomes clear before the final prescription is written in real clinical workflows. We use it to simulate a realistic setting in which the provider has completed diagnostic workup and is now deciding on a treatment plan.\\
To ensure strict temporal consistency and prevent information leakage, we clarify the timestamp alignment between patient input information and the target prescriptions:\\
\textbf{Admission-Time Inputs.} All input data used by our system—such as demographics, chronic conditions, allergies, and social/family history—are derived from structured and semi-structured records available \textbf{at the time of hospital admission}, i.e., before any treatment or prescription is made. This corresponds to the \texttt{admittime} field in the \texttt{mimiciv\_core.admissions} table.\\
\textbf{Prescription Time Filtering.} To avoid utilizing prescriptions influenced by later clinical events, we only retain prescriptions from the \texttt{mimiciv\_hosp.prescriptions} table whose \texttt{starttime} satisfies the following condition:
  \[
  \texttt{starttime} \leq \texttt{admittime} + 24\text{h}
  \]
  This restriction ensures that all target prescriptions fall within the same early-stage window as the admission-time inputs.\\
\textbf{Clinical Assumption.} This setting simulates realistic early-stage clinical decision-making, where medication recommendations are made based solely on admission-available information. It also prevents data leakage and ensures the temporal validity of the prediction task.

Crucially, we do not use any post-prescription information such as the medication list in the discharge summary, outcome sections, or notes written after treatment decisions. Our model inputs include only structured pre-prescription variables (e.g., demographics, allergies, medications on admission) and the diagnosis field mentioned above. The prescription is then generated prospectively, not extracted or reconstructed from final records.

\subsection{Task Details}
The primary task addressed in this study is automated medication recommendation based on a patient's unstructured clinical notes. Each input consists of a clinical note text detailing the patient's medical history, symptoms, diagnoses, lab findings, and prior treatments, written in free-text format. The output is a list of prescribed medications that are clinically appropriate and aligned with established treatment standards. The task requires the model to read and analysis the patient’s conditions, infer relevant clinical risk factors, detect potential risks between medication usage and produce a medication regimen that avoids duplication, contraindications, and inappropriate dosages. This is a multiple rounds generation task with strong emphasis on interpretability, safety, traceability, and aims to pursue consistency with expert prescriptions.
\subsection{Complexity Level}
\begin{figure}[H]
    \centering
    \includegraphics[width=0.8\linewidth]{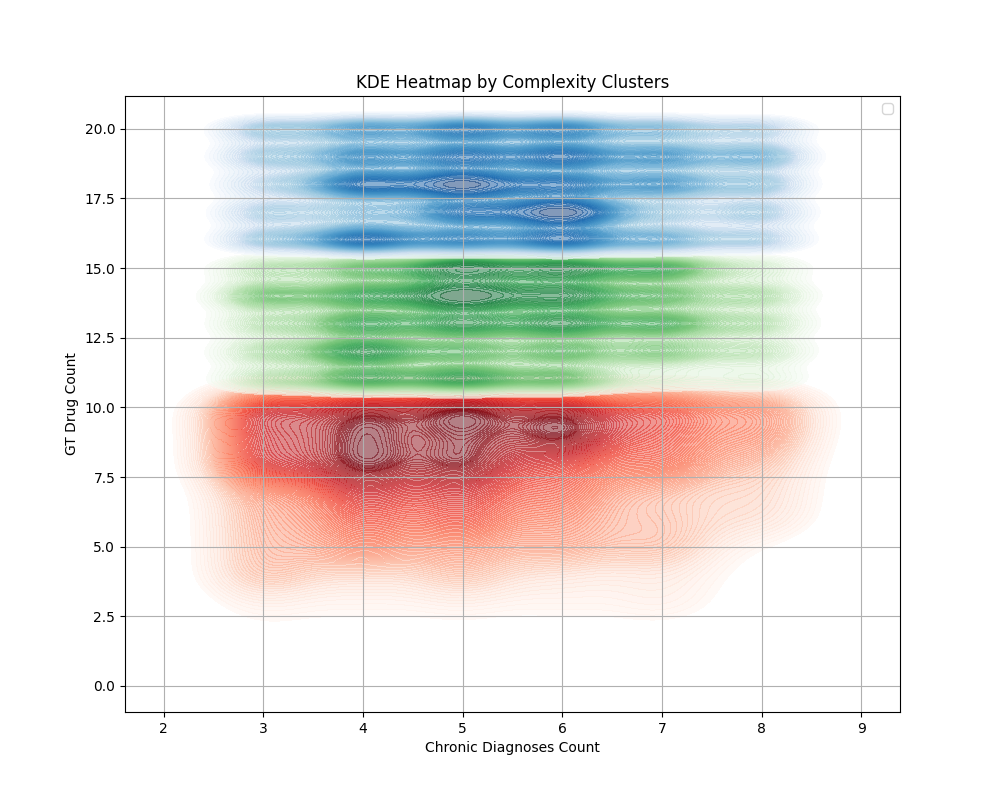}
    \caption{KDE-based Complexity Clusters}
    \label{fig:kde}
    \Description{workflow}
\end{figure}

We define the task complexity level based on two key factors: (1) the number of chronic diseases identified in the patient's clinical records, and (2) the number of medications present in the corresponding ground-truth prescription. A higher number of chronic conditions typically indicates a more complex health profile requiring nuanced clinical judgment. Likewise, longer medication lists increase the risk of drug-drug interactions, contraindications, and dosage conflicts. Therefore, cases with both a high chronic disease count and a large number of prescribed medications are treated as high-complexity scenarios. This stratification enables us to analyze model performance under varying levels of clinical difficulty and assess robustness across simple to complex prescription tasks.
\paragraph{KDE-based Visualization of Complexity Clusters.}
To better illustrate the distribution of patient complexity, we performed a two-dimensional kernel density estimation (KDE) visualization based on the number of chronic diagnoses and the number of distinct drugs (NDC count) shown in Figure ~\ref{fig:kde}.
Patients were first grouped into three complexity levels using \textbf{K-means clustering} ($k=3$). 
For each cluster, we plotted the KDE heatmap with distinct color gradients (\texttt{Reds}, \texttt{Greens}, and \texttt{Blues}) to highlight the density and overlap regions. 
This visualization enables a clear observation of how patient populations with different clinical burdens are distributed across the feature space.

\subsection{DUA compliance checklist}
This study utilized the MIMIC-IV clinical database, which is made available by the MIT Laboratory for Computational Physiology under the PhysioNet credentialed access framework. We confirm that our use of the data fully complies with the requirements stated in the Data Use Agreement (DUA), including but not limited to:
\begin{itemize}
    \item \textbf{Credentialed Access:} All authors accessing the MIMIC-IV data have completed the CITI ``Data or Specimens Only Research'' course and have been granted credentialed access to the dataset through PhysioNet.
    
    \item \textbf{Ethical Use:} The data were used solely for the purposes of academic and non-commercial research. No attempt was made to identify individual patients or institutions within the dataset.
    \item \textbf{Data Security:} The MIMIC-IV data were stored on secure servers and only accessed by credentialed researchers in compliance with institutional data security policies.
    
    \item \textbf{No Re-distribution:} No part of the MIMIC-IV dataset was redistributed or shared with any third parties beyond the scope of credentialed users.
    
    \item \textbf{Privacy and Confidentiality:} All analyses were conducted on de-identified data. No individual-level identifiers were reconstructed or inferred in any form.
    
    \item \textbf{Attribution:} The MIMIC-IV dataset was properly cited and credited in all relevant parts of this work.
\end{itemize}
We acknowledge and respect the efforts of the data custodians and contributors of the MIMIC-IV dataset and ensure that all analyses comply with the ethical and legal standards outlined in the DUA.

\section{Tool Set Details}

\subsection{Programming Language and Core Environment}
We employed a comprehensive suite of tools throughout the course of this study, encompassing all major phases of the research workflow — from data preprocessing and code development to large-scale model deployment, experimental execution, and result visualization. These tools were carefully selected to ensure compatibility, scalability, and efficiency across various tasks. The detailed composition and functionalities of the tools used are summarized in Table~\ref{tab:tooling}.

\begin{table*}[ht]
\raggedright 
\caption{Implementation and Tooling Details}
\label{tab:tooling}
\begin{tabularx}{0.9\textwidth}{@{} lX @{}}
\toprule
\textbf{Component} & \textbf{Tool Usage Description} \\
\midrule
\textbf{Programming Language} & Python 3.12, selected for its mature ecosystem in data science and machine learning. \\
\textbf{Multi-Agent Architecture} & LangChain or CrewAI for agent coordination and tool-agent orchestration. \\
\textbf{Development Environment} & Linux-based environment. \\
\textbf{Ollama Model Deployment} & Ollama platform with local models including general-purpose LLMs and reasoning-optimized LLMs. \\
\textbf{PyTorch} & Used for embedding and handling tokenization. \\
\textbf{RAG} & Retrieval-Augmented Generation with HuggingFace models. \\
\textbf{LangChain} & Framework for multi-agent system orchestration, tool use, and workflow control. \\
\textbf{OpenAI API} & Access to GPT-4o via OpenAI API or local endpoints. \\
\textbf{RxNorm API} & For mapping free-text drug names to RxCUI standardized codes. \\
\textbf{Matplotlib / Seaborn} & Visualization of results, including line plots and heatmaps. \\
\textbf{RxNorm + RxCUI} & For drug normalization and standardization. \\
\textbf{ICD-10} & Classification of chronic conditions from MIMIC-IV \texttt{diagnoses\_icd} tables. \\
\bottomrule
\end{tabularx}
\end{table*}

\section{Baseline Details}
\subsection{Baseline}
To evaluate the effectiveness of our proposed multi-agent framework PRISMATIC, we compared it against several representative prompting-based baseline strategies that have been widely adopted in recent clinical and general NLP reasoning tasks:
\begin{itemize}
    \item \textbf{Zero-Shot Prompting}: The model is directly given a patient clinical note and asked to generate a prescription without any in-context examples or explicit reasoning chains. This baseline tests the model's raw generalization capability.
    \item \textbf{Few-Shot Prompting}: The model receives a few exemplars of (input note → correct prescription) pairs as demonstrations in the prompt. These examples are manually selected to represent common chronic disease profiles and help guide the generation.
    \item \textbf{Chain-of-Thought (CoT)}: This approach encourages the model to perform intermediate reasoning by providing step-by-step thought processes in the prompt. It enhances interpretability and can improve complex decision-making under uncertainty.
    \item \textbf{Tree-of-Thoughts (ToT)}: A structured reasoning method that explores multiple reasoning paths (as a tree structure), evaluates partial thoughts, and backtracks to optimize prescription selection. This method is particularly suited for tasks with high combinatorial complexity like medication recommendation.
    \item \textbf{ReAct (Reasoning and Acting)}: A hybrid prompting framework that combines logical reasoning and knowledge-based actions. In our context, the model reasons about symptoms and then chooses actions (e.g., adding/removing drugs) based on structured feedback.
    \item \textbf{Medprompt}: MedPrompt is a prompting-based framework designed to enhance the performance of large language models (LLMs) on medical question-answering and clinical reasoning tasks. It integrates techniques such as instruction tuning, chain-of-thought prompting, and domain-specific prompt design to guide LLMs toward producing more accurate and explainable medical outputs. By combining retrieval-augmented few-shot examples with consistency-checking strategies, MedPrompt demonstrates that prompt engineering—when tailored to medical contexts—can outperform even fine-tuned models on various benchmarks, while maintaining generalizability across different clinical domains.
\end{itemize}
All baseline methods were tested using the same MIMIC-IV derived input data, and outputs were evaluated using consistent metrics, including Precision, Overlap Rate, and Exact Match Ratio, to ensure fair comparison.

\subsection{Detailed Prompt Examples for Baseline Methods}
The following prompt templates are written for the \textbf{Prescription Generation (PG)} task. 
Each prompt uses the same, richly-specified patient vignette to ensure comparability. 
Prompts include constraints (drug safety, interactions), required output format (structured JSON-like prescription + short rationale), and instructions for evidential justification where applicable.

\bigskip
\noindent\textbf{Patient vignette (shared by all prompts):}
\begin{lstlisting}
Patient: 72-year-old male
Chief complaints: progressive exertional dyspnea for 3 weeks; increased fatigue and confusion over last 2 days.
Past medical history: Type 2 diabetes mellitus (20 years), atrial fibrillation (on warfarin), ischemic heart disease with prior MI (2015), early Alzheimer's disease, chronic kidney disease stage 3 (eGFR 45 mL/min/1.73m^2).
Current medications (reported at admission): 
  - Metformin 500 mg BID
  - Glyburide 5 mg QD
  - Warfarin variable dosing (INR last known 2.1)
  - Simvastatin 40 mg QHS
  - Lisinopril 10 mg QD
Allergies: penicillin (rash)
Family history: brother died suddenly at age 65 (unknown cause)
Social history: former smoker (30 pack-years), quit 10 years ago
Vitals (on admission): BP 140/80 mmHg, HR 96 bpm (irregular), RR 20, SpO2 94% on room air, Temp 36.8°C
Labs (historic / unavailable on admission): no current in-hospital labs available to the model
Clinical constraints: Admission occurs before any new in-hospital labs/imaging.
Goal: Provide an initial safe, pharmacologically rational, and interpretable outpatient/acute medication plan suitable for immediate admission management, minimizing high-risk DDIs and drug–patient contraindications.
\end{lstlisting}

\bigskip
\noindent Below are detailed prompt templates for each baseline. Each prompt expects the output in the following structured format:

\begin{lstlisting}
Output Format (required):
1) Prescription list: [{drug_name, dose, route, frequency, indication, rationale (1-2 sentences), safety_flags}]
2) Short rationale (2-4 sentences) summarizing main safety choices (e.g., DDI avoided, renal adjustments)
3) List of evidence or rules used (e.g., "Avoid NSAIDs with CKD; adjust glyburide in elderly with renal impairment")
\end{lstlisting}

\bigskip
\noindent\textbf{1. Zero-Shot Prompting}
\begin{lstlisting}
Instruction:
You are a clinical decision assistant. Based solely on the patient vignette below and general pharmacology/clinical guidelines, produce an initial admission prescription plan that is safe and interpretable.

Patient vignette:
[PASTE THE SHARED PATIENT VIGNETTE HERE]

Constraints:
- Avoid medications that are high-risk in combination with warfarin (e.g., strong CYP inhibitors or NSAIDs with bleeding risk).
- Adjust or avoid agents that are contraindicated or high-risk in advanced age and CKD stage 3.
- If uncertain about exact dose due to missing labs, choose conservative, lower-end dosing or state that a lab is required before dose escalation.
- Provide explicit safety_flags when a potential issue is present.

Required output: Use the specified Output Format.
\end{lstlisting}

\bigskip
\noindent\textbf{2. Few-Shot Prompting}
\begin{lstlisting}
Instruction:
You are a clinical decision assistant. Use the examples below as demonstrations of ideal prescribing for similar multimorbid older adults, then produce a prescription for the target patient.

--- Example 1 ---
Patient: 68-year-old female with T2DM and osteoarthritis; on warfarin.
Prescription (example): 
- Acetaminophen 500 mg PO q6h prn for pain (avoid NSAIDs due to warfarin and CKD).
- Continue metformin 500 mg BID; hold sulfonylureas if history of hypoglycemia.
Rationale: Avoid NSAIDs on warfarin; choose acetaminophen for analgesia.

--- Example 2 ---
Patient: 75-year-old male, hypertension, mild CKD, on simvastatin.
Prescription (example):
- Continue lisinopril 10 mg QD.
- Switch simvastatin 40 mg to atorvastatin 20 mg due to lower DDI risk if interacting meds are likely.
Rationale: reduce DDI potential with atorvastatin, monitor LFTs.

Target patient:
[PASTE THE SHARED PATIENT VIGNETTE HERE]

Constraints & Output: As described in Output Format. Favor the demonstrated conservative, safety-first choices.
\end{lstlisting}

\bigskip
\noindent\textbf{3. Chain-of-Thought (CoT) Prompting}
\begin{lstlisting}
Instruction:
Reason step-by-step (show your chain-of-thought) before giving the final prescription.
Your reasoning should include: (A) identification of primary problems and competing therapeutic goals, (B) list of high-risk interactions/contraindications to consider, (C) candidate therapeutic options with advantages/disadvantages, and (D) final prescription with explicit safety rationales.

Patient vignette:
[PASTE THE SHARED PATIENT VIGNETTE HERE]

Step-by-step prompt guide:
1) List the top 3 clinical priorities for this patient at admission.
2) For each current medication, evaluate a) DDI risk with warfarin, b) renal dosing concerns for CKD stage 3, c) geriatric safety (e.g., sulfonylurea risk).
3) Propose a conservative medication plan; for each drug indicate dose, route, frequency, and reason.
4) Finally, summarize the prescription in the required Output Format.

Note: The Chain-of-Thought portion will be part of the prompt and used to encourage transparent reasoning; the final prescription must be concise and follow the Output Format.
\end{lstlisting}

\bigskip
\noindent\textbf{4. Tree-of-Thoughts (ToT) Prompting}
\begin{lstlisting}
Instruction:
Explore multiple candidate reasoning paths (a small reasoning tree). For each path, produce a partial prescription and a brief score (0-1) estimating safety and coverage.
Perform up to 3 branching steps, prune low-scoring branches, and return the highest-scoring prescription with rationale.

Patient vignette:
[PASTE THE SHARED PATIENT VIGNETTE HERE]

ToT procedure (you should simulate these steps in your reply):
- Root: consider immediate goals (anticoagulation risk, glycemic control, cardiac/respiratory symptom).
- Branch A: Prioritize minimizing bleeding risk (avoid analgesics that increase bleeding).
- Branch B: Prioritize glycemic safety in elderly with CKD (avoid glyburide).
- Branch C: Prioritize symptom control for dyspnea (short-term symptomatic therapy vs immediate cardiac workup).

For each branch, produce (i) a 1-2 line thought, (ii) a candidate partial prescription, (iii) a safety/coverage score in [0,1].
After pruning, provide the final chosen prescription in the Output Format, plus a short explanation why this tree search selected it.
\end{lstlisting}

\bigskip
\noindent\textbf{5. ReAct (Reasoning + Acting) Prompting}
\begin{lstlisting}
Instruction:
Operate in an interleaved REASON and ACT loop. First, produce a brief reasoning step; then, perform an action (e.g., propose adding/removing/modifying a drug). Repeat up to 4 cycles and converge on a final prescription. Each action must be accompanied by the evidence/rationale.

Patient vignette:
[PASTE THE SHARED PATIENT VIGNETTE HERE]

REASON-ACT sample flow:
[REASON] Identify major safety conflict.
[ACT] Remove drug X / Propose alternative Y with dose and rationale.
[REASON] Evaluate renal dosing for proposed changes.
[ACT] Adjust dose accordingly or request urgent labs.

Produce the final prescription in the Output Format, and include the REASON/ACT log (2-4 cycles) to show decision process.
\end{lstlisting}

\bigskip
\noindent\textbf{6. MedPrompt (Domain-Aware Prompting + RAG + Consistency Checks)}
\begin{lstlisting}
Instruction:
Use domain-aware prompting with retrieval cues (assume a retrieval module that can fetch guideline snippets or drug labels). Perform a stepwise generation with consistency checks:
1) Retrieve: fetch up to 3 evidence snippets relevant to high-risk items (warfarin interactions, CKD dosing, elderly sulfonylurea risk).
2) Propose: generate an initial prescription consistent with retrieved evidence.
3) Verify: run an internal consistency check for DDI, renal dosing, and geriatric safety. If a check fails, revise and repeat up to 2 iterations.
4) Output final prescription with explicit citations (e.g., "evidence: FDA label - warfarin/ibuprofen risk").

Patient vignette:
[PASTE THE SHARED PATIENT VIGNETTE HERE]

Output: Provide the Output Format, plus a short list of the retrieved evidence items (title + short quote or summary).
\end{lstlisting}

\bigskip
\noindent\textbf{Optional: Example Safety-Checking (SC) Prompt Template (for comparison/evaluation agents)}
\begin{lstlisting}
Instruction (Safety Check Agent):
Given a candidate prescription and the patient vignette, identify:
- Any high-severity DDI pairs (with severity label), 
- Any drug–patient incompatibilities (e.g., glyburide in elderly with CKD),
- Any dosage errors requiring urgent lab checks.

Input:
- Patient vignette (as above)
- Candidate prescription (structured list)

Output Format:
1) List of detected issues: [{type: DDI/DPI/Dosage, items: [...], severity: low/moderate/high, evidence}]
2) Overall safety score in [0,1] and brief actionable recommendation (e.g., "hold glyburide and consider insulin sliding scale until glycemic assessment").
\end{lstlisting}

\bigskip
\noindent\textbf{Notes for experimental reproducibility}
\begin{itemize}
  \item Replace the placeholder \texttt{[PASTE THE SHARED PATIENT VIGNETTE HERE]} with the exact vignette block to keep inputs identical across baselines.
  \item For Few-Shot and MedPrompt, the retrieval database and exemplars should be fixed and documented (e.g., which guideline snippets or which exemplar cases).
  \item When comparing methods, measure both (i) clinical safety metrics (DDI counts, DPI flags, dosage violations) and (ii) interpretability metrics (presence of rationale, evidence citations).
\end{itemize}

To ensure fair comparison across all baselines, we use the same patient case as input for the \textbf{Prescription Generation (PG)} task. 
Below we illustrate representative prompt templates used in each baseline setting.

\begin{itemize}
    \item \textbf{Zero-Shot Prompting}
    \begin{lstlisting}[language=Python,frame=single,basicstyle=\ttfamily\footnotesize]
Patient Note:
- Chief Complaint: Persistent cough, shortness of breath
- Medical History: Hypertension, Type 2 diabetes
- Current Medications: Metformin, Amlodipine
- Allergies: None

Task:
Generate a prescription for the above patient based on clinical reasoning.
\end{lstlisting}

    \item \textbf{Few-Shot Prompting}
    \begin{lstlisting}[language=Python,frame=single,basicstyle=\ttfamily\footnotesize]
Example 1:
Patient: Fever, sore throat; History: none.
Prescription: Paracetamol 500mg, 3 times daily.

Example 2:
Patient: Cough, hypertension.
Prescription: Amlodipine 5mg daily, Dextromethorphan syrup.

Now, for the following patient:
- Chief Complaint: Persistent cough, shortness of breath
- History: Hypertension, Type 2 diabetes
Generate a prescription.
\end{lstlisting}

    \item \textbf{Chain-of-Thought (CoT)}
    \begin{lstlisting}[language=Python,frame=single,basicstyle=\ttfamily\footnotesize]
Patient Info:
Cough, shortness of breath; History: Diabetes, Hypertension.

Step 1: Identify possible causes.
Step 2: Review contraindications (e.g., DDI, DPI).
Step 3: Choose suitable medications and dosage.

Let's reason step by step, then provide the prescription.
\end{lstlisting}

    \item \textbf{Tree-of-Thoughts (ToT)}
    \begin{lstlisting}[language=Python,frame=single,basicstyle=\ttfamily\footnotesize]
You are to generate multiple reasoning paths (thoughts):
Thought 1: Treat cough symptomatically.
Thought 2: Consider cardiometabolic comorbidities.
Thought 3: Avoid drug-drug interactions with metformin.

Evaluate each path and select the optimal prescription.
\end{lstlisting}

    \item \textbf{ReAct (Reasoning + Acting)}
    \begin{lstlisting}[language=Python,frame=single,basicstyle=\ttfamily\footnotesize]
[Reasoning] The patient has diabetes and hypertension. 
Avoid steroids; consider ACE inhibitors cautiously.

[Action] Remove contraindicated drug.
[Action] Add expectorant for cough relief.

Final Prescription: ...
\end{lstlisting}

    \item \textbf{MedPrompt}
    \begin{lstlisting}[language=Python,frame=single,basicstyle=\ttfamily\footnotesize]
You are a clinical reasoning model.

Instruction: Generate a safe, effective prescription.
Reasoning: Step-by-step CoT.
Retrieval: Use similar chronic cases from training examples.
Consistency Check: Verify no DDI or contraindication.

Input:
- Cough, shortness of breath; Diabetes, Hypertension.

Output:
Final Prescription + Safety Explanation.
\end{lstlisting}

\end{itemize}

\section{\textbf{\texttt{PRISMATIC}} Details}
\subsection{An example of \textbf{\texttt{PRISMATIC}}}
\begin{figure}[H]
    \centering
    \includegraphics[width=1\linewidth]{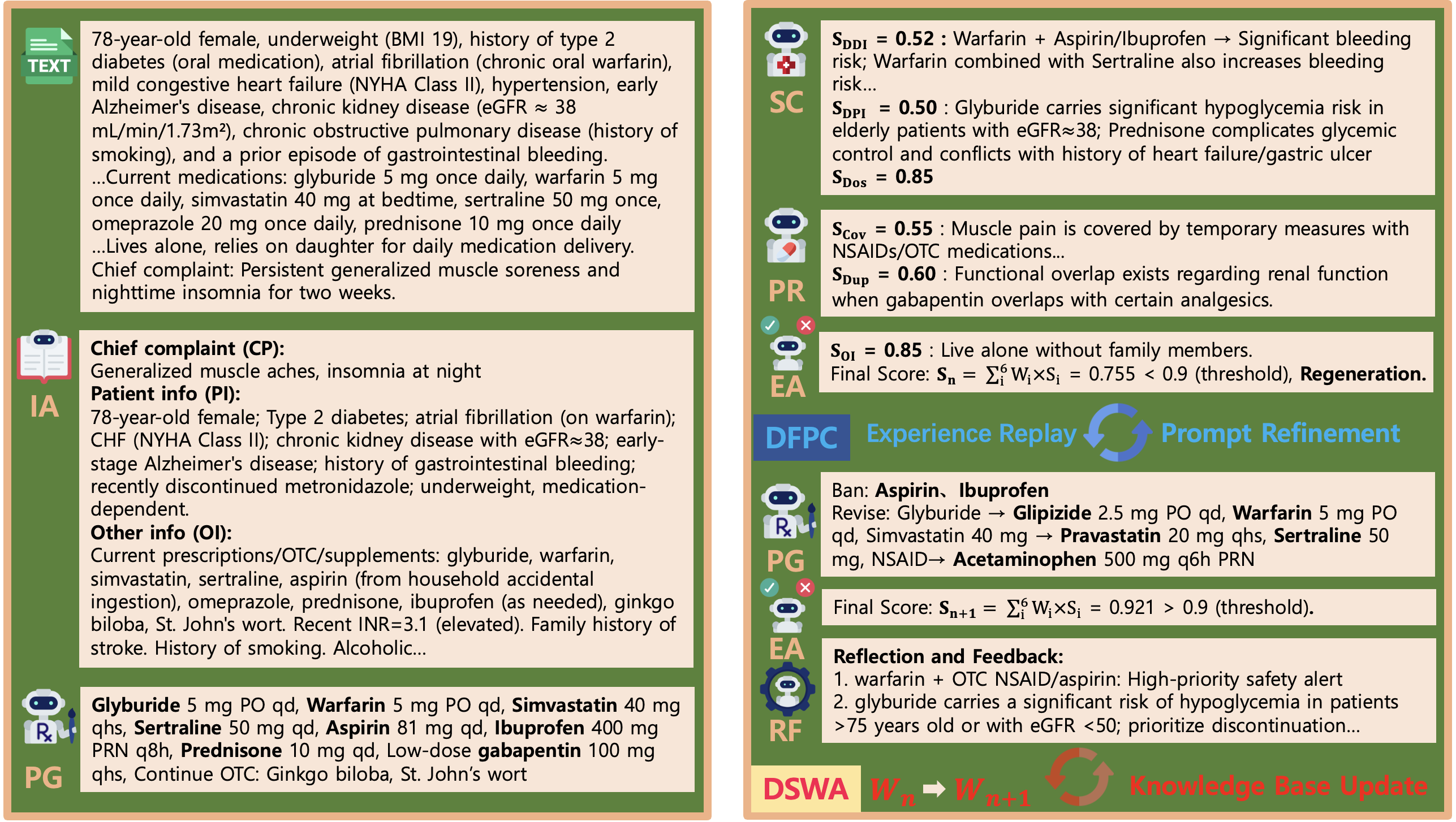}
    \caption{An example of \textbf{\texttt{PRISMATIC}}}
    \label{fig: WorkFlow}
    \Description{workflow}
\end{figure}

\subsection{Formal model of multi-agent collaboration}
We denote the agent society by $\mathcal{A}=\{ \mathrm{IA},\mathrm{PG},\mathrm{SC},\mathrm{PR},\mathrm{EA},\mathrm{RF} \}$.
Let $B_t$ be the \emph{blackboard state} at iteration (round) $t$, which stores structured patient facts,
candidate prescriptions, violation records and provenance links. Each agent $a\in\mathcal{A}$ maintains
a local knowledge base $\mathrm{KB}_a(t)$ and a capability vector $\mathbf{c}_a(t)$ describing its
current competence (e.g., retrieval quality, rule coverage, prompt policy). The system-wide evaluation
weights are $\mathbf{W}(t)=[w_1(t),\dots,w_6(t)]$ controlled by DSWA.

\paragraph{Communication protocol.}
Communication among agents is realized by a shared blackboard $B_t$ and asynchronous message events.
We formalize the blackboard as a tuple
\[
B_t = \big( \mathcal{F}_t, \mathcal{P}_t, \mathcal{L}_t \big),
\]
where $\mathcal{F}_t$ are extracted facts (from IA), $\mathcal{P}_t$ are candidate prescriptions,
and $\mathcal{L}_t$ are structured logs/evidence items. Agents interact by performing atomic operations
on $B_t$:
\[
\texttt{post}(a,\,\phi),\qquad \texttt{read}(a,\,\psi),\qquad \texttt{claim}(a,\,r)
\]
meaning agent $a$ posts assertion $\phi$, reads query $\psi$, or places a temporary claim $r$
(e.g., "I propose to remove drug X"). The protocol supports both asynchronous posting (default) and
synchronous negotiation: an agent may request a synchronous adjudication window $[t,t+\Delta]$ during
which relevant agents must provide responses to a specified claim.

Formally, an asynchronous update step proceeds as:
\[
B_{t+1} \leftarrow B_t \cup \{\texttt{post}(a,\phi)\}\quad\text{(committed atomically)},
\]
whereas a synchronous negotiation for claim $r$ executes a bounded protocol:
\[
\mathrm{CollectResponses}(\{a_i\}, r)\ \xrightarrow{\ \Delta\ } \ \mathrm{AggregateResponse}.
\]

\paragraph{Conflict detection and arbitration.}
Let the Safety Check and Pharmacological agents produce local assessments on a candidate prescription $P$:
\[
\mathrm{SC}(P) \mapsto \mathbf{s}^{\mathrm{SC}}(P),\qquad
\mathrm{PR}(P) \mapsto \mathbf{s}^{\mathrm{PR}}(P),
\]
where each vector contains sub-scores for DDI, DPI, dosage, coverage, duplication, and interpretability.
The Evaluation Agent (EA) aggregates these signals into a global score using DSWA weights:
\[
S(P,t) \;=\; \sum_{i=1}^{6} w_i(t)\, s_i(P).
\]
We adopt the following arbitration policy when agents return conflicting prescriptions or conflicting binary claims.

\begin{enumerate}
  \item \textbf{Severity-precedence rule.} If $\mathrm{SC}$ detects any \emph{high-severity} safety violation
  (labelled by severity function $\sigma(\cdot)$ such that $\sigma\geq\sigma_{\text{high}}$), then the EA
  immediately rejects $P$ irrespective of PR's positive utility, and DFPC must generate targeted corrections:
  \[
  \text{if }\exists\;v\in\mathrm{SC}(P):\ \sigma(v)\ge\sigma_{\text{high}}\quad\Rightarrow\quad S(P,t)\leftarrow 0.
  \]
  This encodes a safety-first constraint propagation step.

  \item \textbf{Weighted consensus rule.} In absence of high-severity violations, EA applies weighted aggregation:
  \[
  \mathrm{Decision}=\begin{cases}
    \text{accept}, & S(P,t)\ge \tau,\\
    \text{invoke DFPC}, & S(P,t)< \tau,
  \end{cases}
  \]
  where threshold $\tau$ is a tunable operational parameter. Under tie or near-tie cases, EA solicits
  a synchronous resolution from SC and PR using evidence-backed RAG snippets; the candidate with higher
  post-evidence score is preferred.

  \item \textbf{Counterfactual verification (optional).} For ambiguous conflicts (e.g., PR prefers drug A,
  SC flags moderate interaction), DFPC can run a bounded counterfactual probe: simulate $P_{\neg d}$ (remove drug $d$)
  and compute $\Delta S = S(P,t)-S(P_{\neg d},t)$. If $\Delta S<0$ (removal increases score) choose $P_{\neg d}$.
\end{enumerate}

The high-level arbitration loop is summarized in pseudo-code (informal):

\begin{algorithm}
\caption{EA arbitration (sketch)}
\begin{flushleft}
\textbf{Input:} candidate $P$, scores $\mathrm{SC}(P),\mathrm{PR}(P)$, weights $\mathbf{W}(t)$\\
\textbf{if} SC reports high-severity violation: reject $P$ and call DFPC \\
\textbf{else} compute $S(P,t)=\sum_i w_i(t)s_i(P)$; \\
\quad \textbf{if} $S(P,t)\ge\tau$ accept $P$; \\
\quad \textbf{else} call DFPC to perform targeted repair.
\end{flushleft}
\end{algorithm}

\section{Main Results}
\subsection{Prompt Engineering Methods}
Figure ~\ref{fig:gpt}, ~\ref{fig:ds}, ~\ref{fig:llama}, \ref{fig:qwen} below present a comparison between the proposed method and \textbf{\texttt{PRISMATIC}} under various contextual prompts. It is evident that \textbf{\texttt{PRISMATIC}} significantly outperforms existing methods in both EM and accuracy.
Within these figures, blue represents complexity level 1, green represents complexity level 2, red represents complexity level 3, and purple indicates the complete dataset.
\begin{figure}[H]
    \centering
    \includegraphics[width=1\linewidth]{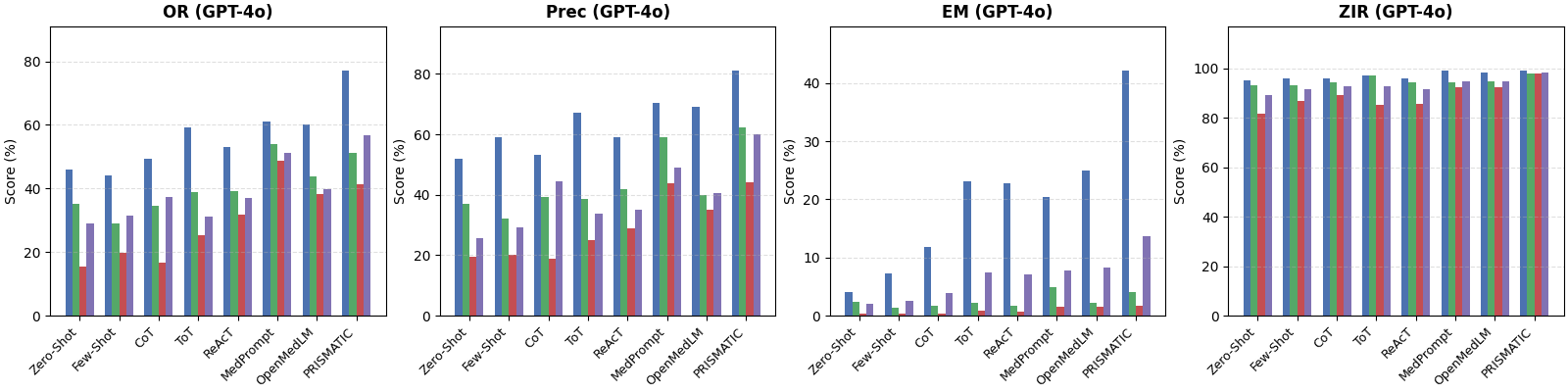}
    \caption{Main Results Comparison with GPT-4o}
    \label{fig:gpt}
    \Description{workflow}
\end{figure}

\begin{figure}[H]
    \centering
    \includegraphics[width=1\linewidth]{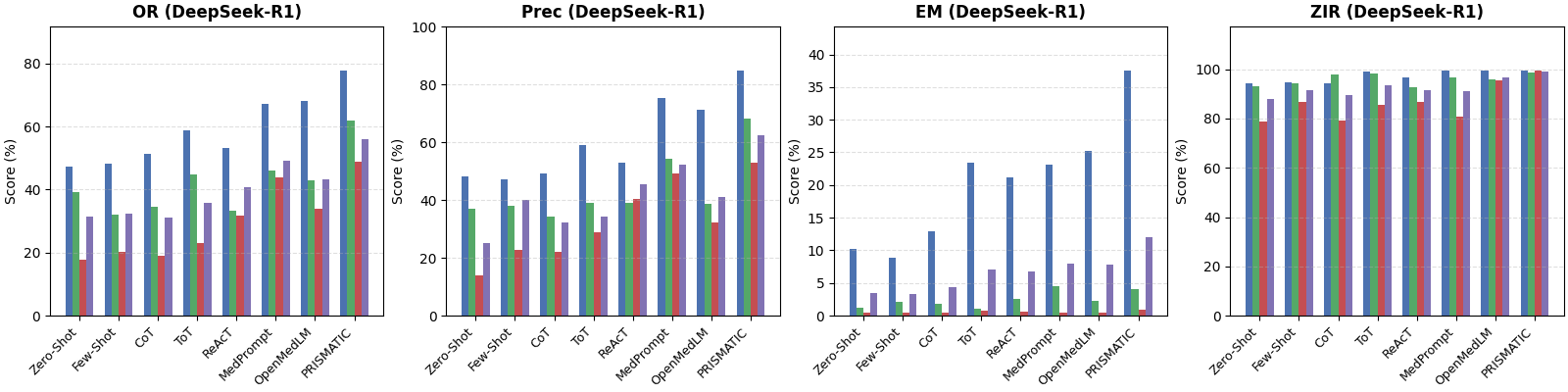}
    \caption{Main Results Comparison with DeepSeek-R1}
    \label{fig:ds}
    \Description{workflow}
\end{figure}

\begin{figure}[H]
    \centering
    \includegraphics[width=1\linewidth]{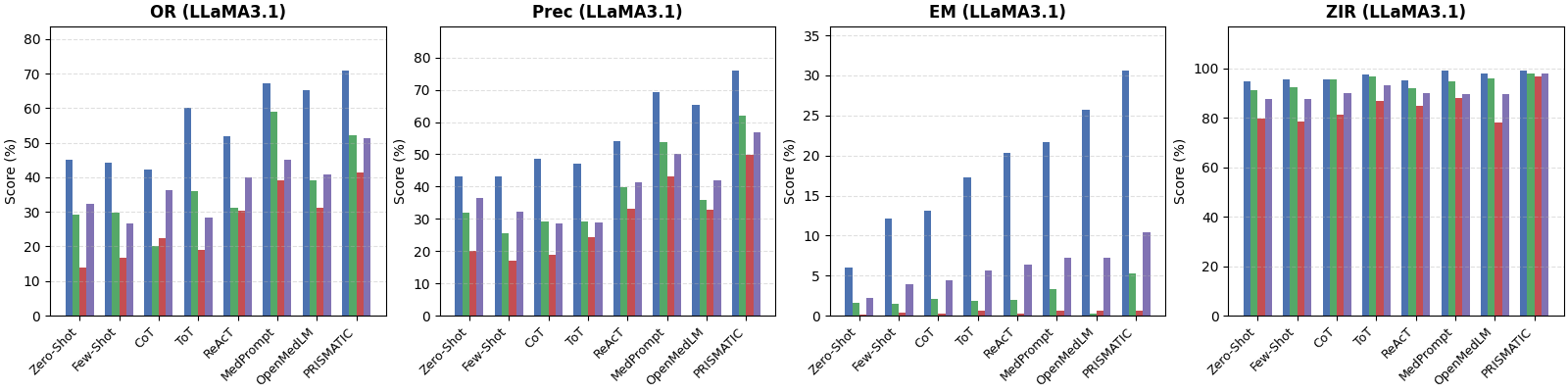}
    \caption{Main Results Comparison with LLama3.1}
    \label{fig:llama}
    \Description{workflow}
\end{figure}

\begin{figure}[H]
    \centering
    \includegraphics[width=1\linewidth]{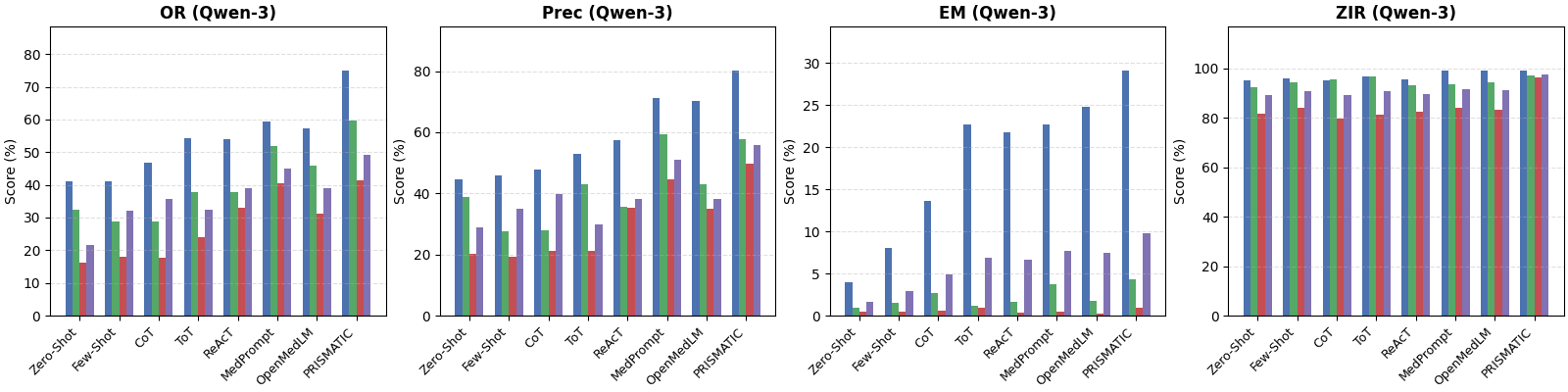}
    \caption{Main Results Comparison with Qwen-3}
    \label{fig:qwen}
    \Description{workflow}
\end{figure}

\subsection{SFT Methods}
\begin{figure}[H]
    \centering
    \includegraphics[width=1\linewidth]{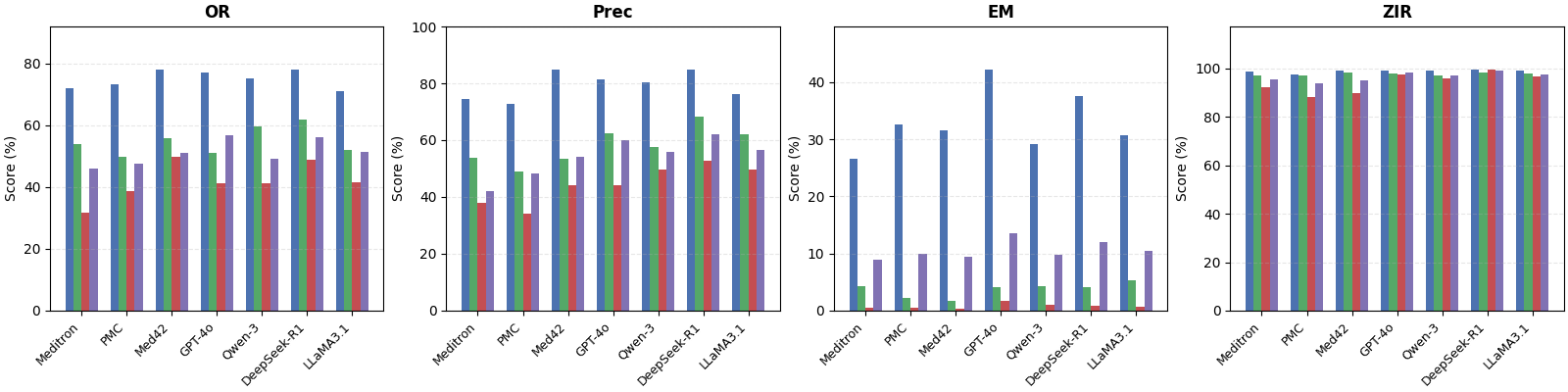}
    \caption{Main Results Comparison with SFT models}
    \label{fig:sft}
    \Description{workflow}
\end{figure}

Figure ~\ref{fig:sft} below compares various fine-tuned models with \textbf{\texttt{PRISMATIC}}. It can be seen that \textbf{\texttt{PRISMATIC}}'s prescription-making capabilities are closer to those of human physicians than those of medical knowledge fine-tuned models.
Within these figures, blue represents complexity level 1, green represents complexity level 2, red represents complexity level 3, and purple indicates the complete dataset.

\section{Additional Implementation Details}
\subsection{Data Preprocessing Details}
We constructed the Chronic Condition Medication (CCM) dataset by preprocessing and integrating multiple tables from the MIMIC-IV database (version 3.1), focusing on patients with multiple chronic conditions and their corresponding medication regimens. The following steps outline the data preprocessing pipeline:
\begin{enumerate}
    \item Patient Selection: We first filtered adult patients (aged ≥ 18) from the patients and admissions tables. To focus on chronic care, we included only those with at least three documented chronic conditions (e.g., hypertension, diabetes, chronic kidney disease) based on ICD-10 codes from the \(diagnoses_{icd}\) table.
    \item Diagnosis Filtering: Using standard chronic disease ICD-10 groupings (e.g., CCS, CMS CCW), we mapped diagnosis codes to a curated list of 15 common chronic diseases. Patients were grouped by the number of distinct chronic conditions they had (e.g., 3–8).
    \item Medication Extraction: Medication orders were extracted from the prescriptions table. We filtered out non-oral routes (e.g., topical, ophthalmic) and retained only structured drug entries with valid RxNorm Concept Unique Identifiers (RxCUIs). Duplicate entries and PRN (as needed) medications were excluded to ensure consistency.
    \item Note Matching: We aligned the medications with their corresponding clinical notes from the noteevents and discharge sections in mimic-iv-note. Each patient’s diagnosis summary and hospital discharge note served as input to our model.
    \item Prescription Cleaning: For each patient admission, we constructed a unique ground-truth prescription by merging all eligible medications during that stay, removing dose/form variations (via RxNorm normalization), and de-duplicating the final list.
    \item Task Stratification: We stratified patient cases by task complexity, using the number of chronic diseases and prescription size as joint indicators. This allowed for performance benchmarking across varying levels of clinical difficulty.
\end{enumerate}
The resulting CCM dataset contains X,XXX samples across a wide range of chronic profiles and prescription combinations, providing a realistic and challenging benchmark for prescription recommendation systems.

\subsection{Evaluation Metrics Details}
To comprehensively assess the effectiveness of our proposed prescription generation framework, we employ three widely recognized evaluation metrics from clinical decision support and medical NLP literature. These metrics capture precision, coverage, and holistic correctness of the generated medication plans:
\begin{itemize}
    \item \textbf{Precision (Prec.)}: Precision measures the proportion of predicted medications that are correct, i.e., found in the gold-standard prescription. It reflects the system's ability to avoid recommending irrelevant or harmful drugs, a crucial requirement in clinical settings. This metric is commonly used in medical recommendation systems to assess the relevance of recommendations.
    \item \textbf{Overlap Rate (OR)}: Overlap Rate evaluates the proportion of gold-standard medications that are successfully retrieved. Unlike precision, this metric emphasizes coverage, ensuring that key drugs are not omitted. It is particularly useful when missing a medication could lead to patient harm, and has been used in prior clinical text mining works.
    \item \textbf{Exact Match Ratio (EM)}: Exact Match Ratio assesses the strict correctness of the predicted prescription by computing the percentage of predictions that exactly match the full gold-standard drug list.
    \item \textbf{Zero Interaction Ratio (ZIR)}: ZIR measures the percentage of generated prescriptions that are completely free of known drug-drug interactions. This metric directly evaluates the system's ability to satisfy a primary clinical objective.
\end{itemize}

\subsection{Multi-agent Details}
In the subsequent subsections, we detail the prompt templates employed in \textbf{\texttt{PRISMATIC}}. The complete version of the prompts is available at our code repository due to space limitations. The following is a description of the different agent prompt templates according to the three levels in turn.

\vspace{4pt}
\begin{mdframed}[linecolor=black, backgroundcolor=gray!10!white, roundcorner=4pt, linewidth=1pt]
\textbf{Information Analysis Agent (IA)}
\hfill
\raisebox{-0.2em}{\includegraphics[width=1.6em,height=1.6em]{Agents/IA.png}}

\vspace{4pt}
\hrule
\vspace{4pt}

\textbf{Input:} Clinical Free Statement Text \textbf{T}

\textbf{Output:} Patient Classification Information \textbf{\{CP, PI, OI\}}

\textbf{Function:} Parses unstructured patient clinical notes into structured components, 
including chief complaints (\textbf{CP}), patient information (\textbf{PI}), and other information (\textbf{OI}).

\textbf{Responsibility:} 
Transforms unstructured text into structured classification sequences to prevent prescription omissions or misclassifications caused by incomplete information extraction and improper categorization.
\end{mdframed}

\vspace{6pt}

\begin{mdframed}[linecolor=black, backgroundcolor=gray!10!white, roundcorner=4pt, linewidth=1pt]
\textbf{Prescription Generation Agent (PG)}
\hfill
\raisebox{-0.2em}{\includegraphics[width=1.6em,height=1.6em]{Agents/PG.png}}

\vspace{4pt}
\hrule
\vspace{4pt}

\textbf{Input:} 
\begin{itemize}
    \item Structured Patient Information \textbf{\{CP, PI, OI\}}
    \item Previous Prescription $\textbf{P}_n$ (if exist)
\end{itemize}

\textbf{Output:} Draft $\textbf{P}$ or Revised Prescription $\textbf{P}_{n+1}$

\textbf{Function:} Analyzes structured patient data from the blackboard to propose a prescription candidate. 

\textbf{Responsibility:} 
\begin{itemize}
    \item Serves as the primary operator of the inner loop, producing draft regimens $\textbf{P}_n$ that form the baseline for downstream safety, pharmacological, and interpretability evaluations.
    \item If the prescription $\textbf{P}_n$ fails to meet the safety standards for output, it adjusts the prescription and outputs an improved $\textbf{P}_{n+1}$ based on feedback.
\end{itemize}
\end{mdframed}

\vspace{6pt}

\begin{mdframed}[linecolor=black, backgroundcolor=gray!10!white, roundcorner=4pt, linewidth=1pt]
\textbf{Safety Check Agent (SC)}
\hfill
\raisebox{-0.2em}{\includegraphics[width=1.6em,height=1.6em]{Agents/SC.png}}

\vspace{4pt}
\hrule
\vspace{4pt}

\textbf{Input:} Prescription $\textbf{P}_n$

\textbf{Output:} Safety Scores
\begin{itemize}
    \item DDI Score: $\textbf{S}_{\textbf{DDI}}$
    \item DPI Score: $\textbf{S}_{\textbf{DPI}}$
    \item Dosage Score: $\textbf{S}_{\textbf{Dos}}$
\end{itemize}

\textbf{Function:} Examines drug–drug interactions (DDI), drug–patient interactions (DPI), and dosage compliance against medical standards. SC employs retrieval-augmented generation (RAG) to conduct safety analyses based on drug package inserts.

\textbf{Implementation:}
\begin{itemize}
    \item $\textbf{S}_{\textbf{DDI}} = 1 - \frac{\#\text{detected DDI pairs}}{\#\text{all possible pairs}}$
    \item $\textbf{S}_{\textbf{DPI}} = 1 - \frac{\#\text{contraindicated drugs}}{\#\text{prescribed drugs}}$
    \item $\textbf{S}_{\textbf{Dos}} = \frac{\#\text{drugs within safe dosage}}{\#\text{total drugs}}$
\end{itemize}

\textbf{Responsibility:} Ensures prescriptions do not contain dangerous combinations or inappropriate dosages, serving as the primary safety safeguard against adverse drug events.
\end{mdframed}

\vspace{6pt}

\begin{mdframed}[linecolor=black, backgroundcolor=gray!10!white, roundcorner=4pt, linewidth=1pt]
\textbf{Pharmacological Rationality Agent (PR)}
\hfill
\raisebox{-0.2em}{\includegraphics[width=1.6em,height=1.6em]{Agents/PR.png}}

\vspace{4pt}
\hrule
\vspace{4pt}

\textbf{Input:} Prescription $\textbf{P}_n$, Chief Complaint ($\textbf{CP}$)

\textbf{Output:} Pharmacological Rationality Scores
\begin{itemize}
    \item Coverage Score $\textbf{S}_{\textbf{Cov}}$
    \item Duplication Score $\textbf{S}_{\textbf{Dup}}$
\end{itemize}

\textbf{Function:} Evaluates prescription \textit{coverage} and \textit{duplication} by checking whether all chief complaints are adequately addressed and redundant drugs avoided.

\textbf{Implementation:}
\begin{itemize}
    \item $\textbf{S}_{\textbf{Cov}} = \frac{\#\text{treated conditions}}{\#\text{total conditions}}$
    \item $\textbf{S}_{\textbf{Dup}} = 1 - \frac{\#\text{redundant drugs}}{\#\text{total drugs}}$
\end{itemize}

\textbf{Responsibility:} Ensures pharmacological comprehensiveness and avoids redundancy, improving clinical efficacy and reducing patient burden.
\end{mdframed}

\vspace{6pt}

\begin{mdframed}[linecolor=black, backgroundcolor=gray!10!white, roundcorner=4pt, linewidth=1pt]
\textbf{Evaluation Agent (EA)}
\hfill
\raisebox{-0.2em}{\includegraphics[width=1.6em,height=1.6em]{Agents/EA.png}}

\vspace{4pt}
\hrule
\vspace{4pt}

\textbf{Input:} Prescription $\textbf{P}_{n}$

\textbf{Output:} 
\begin{itemize}
    \item Other Information Score $\textbf{S}_\textbf{OI}$
    \item Final Evaluation Score $\textbf{S}_n$
\end{itemize}

\textbf{Function:} 
Analyzes omissions, conflicts, and contextual risks using \textbf{OI} extracted by \textbf{IA}. 
The sub-score $\textbf{S}_{\textbf{OI}} \in [0,1]$ is produced by an interpretability-aware reasoning LLM. 
Then, combines this with the outputs of \textbf{SC} and \textbf{PR} to compute the final score $\textbf{S}_n$.

\textbf{Responsibility:} 
Acts as the system’s decision gatekeeper. 
If $\textbf{S}_n < \tau$, triggers \textbf{DFPC} to identify discrepancies and re-initiate correction.

\textbf{Final Score:}
\[
\textbf{S}(t) = \sum_{i=1}^{6} w_i(t)\, \textbf{S}_i(t)
\]

\textbf{Rule:}
\begin{itemize}
    \item If $\textbf{S}(t) \geq \tau$, prescription is accepted.
    \item Else, DFPC is triggered for re-evaluation.
\end{itemize}
\end{mdframed}

\vspace{6pt}

\begin{mdframed}[linecolor=black, backgroundcolor=gray!10!white, roundcorner=4pt, linewidth=1pt]
\textbf{Reflection \& Feedback Agent (RF)}
\hfill
\raisebox{-0.2em}{\includegraphics[width=1.6em,height=1.6em]{Agents/RF.png}}

\vspace{4pt}
\hrule
\vspace{4pt}

\textbf{Input:} Generated Prescription $\textbf{P}_\text{final}$, Gold-standard Prescription $\textbf{P}_\text{gold}$

\textbf{Output:} Knowledge Base Updates $\textbf{KB}$

\textbf{Responsibility:} 
Compares the two prescriptions and summarizes discrepancies into updates for each agent’s knowledge base. 
This mechanism enables the system to continually align with evolving clinical standards, 
promoting reliability and domain adaptation over time.
\end{mdframed}

\subsection{\includegraphics[width=1.6em,height=1.6em]{Mechanism/dswa.png} Inner-loop: Short-term Discrepancy Feedback Precise Calibration (DFPC)}

\begin{itemize}
\item \textbf{Input:} Candidate prescription $\textbf{P}_n$, Evaluation feedback $\textbf{S}_n$ from collaborating agents
\item \textbf{Mechanistic Role:} \textbf{DFPC} functions as an adaptive inner-loop negotiation process that enables agents to locally reconcile conflicts and refine their outputs based on precise, evidence-driven feedback. Rather than regenerating entire prescriptions from scratch, DFPC decomposes the evaluation feedback into interpretable discrepancy signals and redistributes them across the relevant sub-agents, triggering targeted calibration at the level of dosage, contraindications, or coverage.

\item \textbf{Mechanism:}
\begin{itemize}
    \item \textbf{Error Attribution and Local Calibration.} \\
    When $\textbf{S}_n$ falls below the acceptance threshold, DFPC identifies high-impact deviations by tracing them in the shared blackboard memory. Instead of treating the error as global, DFPC attributes it to local causal sources—such as drug–drug conflicts or under-covered indications—and selectively re-engages the corresponding functional agents. This localized correction process reflects the principle of decentralized optimization common in multi-agent reinforcement systems.

    \item \textbf{Experience-grounded Replay.} \\
    The identified errors are transformed into structured feedback tuples $(c_i, \delta_i, e_i)$ capturing condition, deviation, and evidence. These tuples are replayed into the agent’s knowledge buffer, functioning as a lightweight self-supervised signal that helps each agent recall optimal prior decisions under similar clinical contexts.

    \item \textbf{Prompt Adaptation and Reinforcement.} \\
    Following each refinement cycle, the involved agents adapt their local prompting schema, embedding the new contextual constraints into future reasoning traces. Through this iterative calibration, the system demonstrates a form of bounded rationality—balancing correction accuracy with computational efficiency.
\end{itemize}

\item \textbf{System-level Function:}

Conceptually, DFPC serves as a \emph{short-term cognitive correction loop}, analogous to reflexive adaptation in distributed cognition. It minimizes intra-round divergence and accelerates convergence across agents, ensuring that multi-agent deliberation remains stable and clinically aligned without external supervision.
\end{itemize}

\subsection{\includegraphics[width=1.6em,height=1.6em]{Mechanism/dfpc.png} Outer Loop: Long-term Dynamic Self-updating Weight Adjustment (DSWA)}
\begin{itemize}
\item \textbf{Input:} Final prescription $\textbf{P}{\text{final}}$, gold-standard reference $\textbf{P}{\text{gold}}$, and current evaluation weights $\textbf{W}n$
\item \textbf{Output:} Updated weight distribution $\textbf{W}{n+1}$
\item \textbf{Mechanistic Role:}

While DFPC focuses on short-term reflexive correction, \textbf{DSWA} operates at the meta-learning level, governing the system’s long-term adaptation across cases. It dynamically adjusts the internal evaluation landscape by reallocating weights among the six clinical risk dimensions—drug–drug interaction (DDI), drug–patient incompatibility (DPI), dosage error (DOS), coverage deficiency (COV), duplication (DUP), and overall interpretability (OI).

\item \textbf{Mechanism:}
\begin{itemize}
    \item \textbf{Adaptive Weight Evolution.} \\
    After each complete prescription cycle, DSWA quantifies the discrepancy vector between $\textbf{P}_{\text{final}}$ and $\textbf{P}_{\text{gold}}$, and updates the weight vector $\textbf{W}$ according to the magnitude and recurrence of errors across dimensions. Dimensions with persistent or high-severity discrepancies are upweighted, signaling future agents to allocate more reasoning capacity to those safety-critical aspects. This adaptive reweighting mechanism establishes a feedback loop analogous to multi-agent policy shaping under evolving task priorities.

    \item \textbf{Distributed Knowledge Consolidation.} \\
    Each update triggers a synchronization event across agents’ local knowledge bases. Discrepancy logs are distilled into rule-like update entries and injected back into the global repository, gradually aligning distributed memory representations. Over time, this mechanism supports continual learning and cross-agent generalization, reducing redundancy and reinforcing cooperative consistency across the system.

    \item \textbf{Emergent Behavior.} \\
    Through repeated DSWA updates, the system exhibits emergent prioritization behavior: dimensions related to dosage and coverage naturally gain higher long-term weights, reflecting their higher error sensitivity in multimorbid patient populations. This emergent emphasis mirrors adaptive specialization commonly observed in reinforcement-based multi-agent systems.
\end{itemize}

\item \textbf{System-level Function:}

DSWA serves as the \emph{strategic adaptation layer} of PRISMATIC, enabling long-horizon optimization through cumulative self-correction. Together with DFPC, it establishes a two-level learning architecture—where inner-loop agents perform reflexive adjustments and the outer-loop controller refines the collective evaluation policy—thereby bridging immediate reaction and long-term stability within multi-agent clinical decision-making.
\end{itemize}

\section{Limitations}
This study has several limitations that define its scope and suggest future research directions:
\begin{itemize}
    \item \textbf{Incomplete Clinical Scope:} The ground truth prescriptions used as references often include preemptive, supportive, or prognostic medications that address comorbidities, complication prevention, or long-term patient management. These prescriptions reflect complex clinical judgments extending beyond the primary diagnosis. However, our system primarily focuses on generating prescriptions directly related to the diagnosed condition, which may omit such broader therapeutic considerations routinely made by clinicians. This gap limits the system’s ability to fully capture the holistic medication strategies used in real-world practice.
    \item \textbf{Limited Prescription Accuracy and Generalizability:} Although the system incorporates advanced mechanisms such as dynamic self-updating weights and differential feedback calibration to iteratively improve performance, the overall accuracy and alignment with expert prescriptions remain suboptimal, especially in complex cases involving multiple conditions and medications. The prescription generator currently struggles to precisely select optimal drugs, dosages, and administration routes in diverse clinical scenarios. Moreover, the system’s performance is constrained by the scope and richness of the medical knowledge integrated. Enhancing domain coverage with more comprehensive clinical guidelines, drug databases, and real-world practice patterns is necessary to increase robustness and clinical applicability.
    \item \textbf{Evaluation Constraints:} Our evaluation relies heavily on retrospective datasets and reference prescriptions, which may not fully represent real-time clinical decision-making dynamics or patient-specific nuances. The absence of prospective validation in live clinical settings restricts our ability to assess the system’s practical utility and safety in everyday healthcare environments.
\end{itemize}